\documentclass{clv3}

\usepackage[utf8]{inputenc}
\usepackage[T1]{fontenc}
\usepackage{hyperref}
\usepackage{balance}
\usepackage{xcolor}
\usepackage{booktabs}
\usepackage{lscape}
\usepackage{paralist}
\usepackage{chngpage}
\usepackage{tablefootnote}
\definecolor{darkblue}{rgb}{0, 0, 0.5}
\hypersetup{colorlinks=true,citecolor=darkblue, linkcolor=darkblue, urlcolor=darkblue}
	\newcommand\eg{{e.g.,\ }}
	\newcommand\Eg{{E.g.,\ }}
        \usepackage{verbatim}
\usepackage{anyfontsize}
\usepackage[normalem]{ulem}

 \usepackage{marginnote}

\issue{1}{1}{2018}

\runningtitle{Survey of Computational Approaches to Lexical Semantic Change}

\runningauthor{Nina Tahmasebi \emph{et al.}}

\makeatletter
\newcommand\tableofcontents{%
	 \section*{\contentsname}
	 \@starttoc{toc}
	 \par\vspace{13\p@}}
\renewcommand\normalsize{%
   \@setfontsize\normalsize\@ixpt{11}%
   \abovedisplayskip 11\p@ \@plus2\p@
   \belowdisplayskip      \abovedisplayskip
   \abovedisplayshortskip \abovedisplayskip
   \belowdisplayshortskip \abovedisplayskip
   \let\@listi\@listI}
\makeatother

\title{Survey of Computational Approaches to Lexical Semantic Change Detection}
\author{Nina Tahmasebi\thanks{Box 200, SE-405 30 Gothenburg, Sweden, E-mail: nina.tahmasebi@gu.se}}
\affil{University of Gothenburg}

\author{Lars Borin\thanks{Box 200, SE-405 30 Gothenburg, Sweden, E-mail: lars.borin@svenska.gu.se}}
\affil{University of Gothenburg}

\author{Adam Jatowt\thanks{Yoshida-honmachi, Sakyo-ku, Kyoto 606-8501 Japan, E-mail: adam@dl.kuis.kyoto-u.ac.jp}}
\affil{Kyoto University}

\begin{document}
\maketitle

\begin{abstract}

Our languages are in constant flux driven by external factors such as cultural, societal
and technological changes, as well as by only partially understood internal motivations. Words acquire new meanings and
lose old senses, new words are coined or borrowed from other languages
and obsolete words slide into obscurity. Understanding the
characteristics of shifts in the meaning and in the use of words is useful for those who work with the content of historical
texts, the interested general public, but also in and of itself. 

The findings from automatic lexical semantic change detection, and the models of diachronic conceptual change are also currently being incorporated in approaches for measuring document across-time similarity, information retrieval from long-term document archives, the design of OCR algorithms, and so on. In recent years we have seen a surge in interest in the academic community in computational methods and tools supporting inquiry into diachronic conceptual change and lexical replacement.
This article provides a comprehensive survey of recent computational techniques to tackle both.

\end{abstract}

\section{Introduction}\label{sec:intro}
Vocabulary change has long been a topic of interest to linguists and the general public alike. This is not surprising  considering the central role of language in all human spheres of activity, together with the fact that words are its most salient elements. Thus it is natural that we want to know the ``stories of the words we use'' including when and how words came to possess the senses they currently have as well as what currently unused senses they had in the past. Professionals and the general public are interested in the origins and the history of our language as testified to by numerous books on semantic change aimed at a wide readership. 

Traditionally, vocabulary change has been studied by linguists and other scholars in the humanities and social sciences with manual, ``close-reading'' approaches. While this is still largely the case inside linguistics, recently we have seen proposals, originating primarily from computational linguistics and computer science, for how semi-automatic and automatic methods could be used to scale up and enhance this research.

Indeed, over the last two decades we have observed a surge of research papers dealing with detection of lexical semantic changes and formulation of generalizations about them, based on datasets spanning decades or centuries. With the digitization of historical documents going on apace in many different contexts, accounting for vocabulary change has also become a concern in the design of information access systems for this rapidly growing body of texts. At the same time, as a result, large scale corpora are available allowing the testing of computational approaches for related tasks and providing quantitative support to proposals of various hypotheses.

Despite the recent increase in research using computational approaches to investigate lexical semantic changes, the community lacks a solid and extensive overview of this growing field. The aim of the present survey is to fill this gap. While we were preparing this survey article, two related surveys appeared, illustrating the timeliness of  the topic. The survey by \citet{kutuzov-etal-2018} has a narrower scope, focusing entirely on diachronic word embeddings. The broader survey presented by \citet{tang-2018} covers much of the same field as ours in terms of computational linguistics work, but provides considerably less discussion of the connections and relevance of this work to linguistic research. A clear aim in preparing our presentation has been to anchor it firmly in mainstream historical linguistics and lexical typology, the two linguistics subdisciplines most relevant to our survey. Further, the application of computational methods to the study of language change has gained popularity in recent years. Relevant work can be found not only in traditional linguistics venues, but can be found in journals and conference proceedings representing a surprising variety of disciplines, even outside the humanities and social sciences. Consequently, another central aim of this survey has been to provide pointers into this body of research, which often utilizes datasets and applies methods originating in computational linguistics research. Finally, our main concern here is with computational linguistic studies of vocabulary change utilizing empirical diachronic (corpus) data. We have not attempted to survey a notable and relevant complementary strand of computational work aiming to simulate historical processes in language, including lexical change (see \citealp{baker-2008} for an overview).

The work surveyed here falls into two broad categories. One is the modeling and study of \emph{diachronic conceptual change} (i.e., how the meanings of words change in a language over shorter or longer time spans). This strand of computational linguistic research is closely connected to corresponding efforts in linguistics, often referring to them and suggesting new insights based on large-scale computational studies, (\eg in the form of ``laws of semantic change''). This work is surveyed in Section~\ref{sec:wsch}, and split into one section on word-level change in Section~\ref{sec:WLSC}, and one on sense-differentiated change in Section~\ref{sec:sdcd}. The word-level change detection considers both count-based context methods as well as those based on neural embeddings, while the sense-differentiated change covers topic modeling-based methods, clustering-based, and word sense induction-based methods.  

The other strand of work focuses on \emph{lexical replacement}, where different words express the same meaning over time. This is not traditionally a specific field in linguistics, but it presents obvious complications for access to historical text archives, where relevant information may be retrievable only through an obsolete label for an entity or phenomenon. This body of work is presented in Section~\ref{sec:dwr}. 

The terminology and conceptual apparatus used in works on lexical semantic change are multifarious and not consistent over different fields or often even within the same discipline. For this reason, we provide a brief background description of relevant linguistic work
 in Section~\ref{sec:lingappr}.

Much current work in computational linguistics depends crucially on (formal, automatic, quantitative, and reproducible) \emph{evaluation}. Given the different aims of the surveyed research, evaluation procedures will look correspondingly different. We devote Section~\ref{sec:method} to a discussion of general methodological issues and evaluation.

A characteristic feature of computational linguistics, is the close connection to concrete computer applications. In Section~\ref{sec:appl} we describe a range of relevant applications presented in the literature, both more research-oriented (\eg visualizations of vocabulary change) and downstream applications, typically information access systems.

We end with a summary of the main points garnered from our literature survey, and provide a conclusion and some recommendations for future work (Section~\ref{sec:summary}).

We believe our survey can be helpful for both researchers already working on related topics as well as for those new to this field, for example, for PhD candidates who wish to quickly grasp the recent advances in the field and pinpoint promising research opportunities and directions.

\section{Conceptual Framework and Application}\label{sec:framework}

\subsection{Model and Classification of Language Change}

Lexical change can be seen as a special case of lexical variation, which can be attributable to many different linguistic and extralinguistic factors. In other words, we see the task of establishing that we are dealing with variants of the same item (in some relevant sense) -- items of form or content -- as logically separate from establishing that the variation is classifiable as lexical change.

However, in language, form and function are always interdependent (this assumption defines
the core of the Saussurean linguistic sign), so in actual practice, we
cannot recognize linguistic forms in the absence of their meanings and vice
versa. This principle is not invalidated by the fact that many
orthographies provide (partial) analyses of the form--meaning
complexes, in the form of word spacing or different word initial, final and medial shapes of letters, etc. There are still many cases where orthographic conventions
do not provide enough clues, a notable example being multiword
expressions, and in such cases tokenization simply does not provide
all and only the lexical forms present in the text
\citep{dridan-oepen-2012}.

Investigation of lexical change is further complicated by the fact that -- as just noted -- observed variation in lexical form between different text materials need not be due to diachronic causes at all, even if the materials happen to be from different time periods. Linguists are well aware that even seen as a synchronic entity, language is full of variation at all linguistic levels. In
spoken language, this kind of variation is the norm. Words have a wide
range of pronunciations depending on such factors as speech rate,
register/degree of formality, phonetic and phonological context, etc. If the
language has a written form, some of this variation may be reflected
in the orthography, but orthography may also reflect ambiguous
principles for rendering some sounds in writing, as when /s/ can be
written alternatively with <s>, <c> and <z> in Swedish. Spelling principles -- if standardized at all, which often is not the case in older texts -- may
change over time independently of any changes in pronunciation
(``spelling reforms''), and in such situations written texts may
exhibit a mix of the older and newer orthography. Finally, in many modern text types we find a large number of spellings which deviate from the standard orthography \citep{eisenstein-2015}.

A central concern of historical linguistics has long been sound
change, which can have a complicated relation to the orthography of a
language. At one point, spoken English allowed the initial consonant
cluster /kn-/, but with time regular sound change reduced this to
/n-/. The cluster is still present in the orthography, however:
\emph{knead, knit} are pronounced exactly the same as \emph{need,
  nit}.

Given enough time, the cumulative effect of regular sound changes can
result in cognate items showing no common sounds at all, \eg Polish
\emph{w} and Swedish \emph{i}, both meaning `in' (preposition) and
both going back to the same (reconstructed) Proto-Indo-European item
$^*\text{\emph{\d{n}}}$, also meaning `in'. There is a substantial
body of work in computational linguistics on developing methods for
``rolling back'' sound changes so that cognates can be identified
among related languages, using both word lists and parallel corpora
\citep{borin-saxena-2013,list-etal-2017}. This is a difficult problem
in the general case. Fortunately, much of the work surveyed here
focuses on much shorter time spans than the 6 millennia or so
separating Polish and Swedish from their reconstructed common Proto-Indo-European
ancestor language, and it furthermore deals exclusively with
written texts. Consequently, and even though we in principle may need
to reckon with some sound changes happening even in such short periods,
form variation in this material can be thought of as \emph{spelling
variation} (\emph{spelling change} if diachronic), and this is how we will refer to it.

A fundamental question underlying all work on semantic change is the
problem of identifying like with like, or -- on the form side --
classifying text words under relevant lexical units, and -- on the
content side -- identifying and grouping together relevant
senses.

Although often trivial (in many writing systems), even the
former task is complicated by the existence of multiword expressions,
the need for word segmentation (in speech and some writing systems),
and -- \emph{a fortiori} in a diachronic context -- language
variation, which may be purely orthographic, both synchronically and
diachronically, as well as a reflection of sound change in the
diachronic setting.

The latter task is widely recognized to be unsolved, and possibly
not even amenable to finding one solution in that there will not be one canonical sense set for a
particular language, but several sets depending both on their intended
use \citep{kilgarriff-1998}, on particular analytical traditions
(``lumpers'' vs. ``splitters''), and even on individual
idiosyncrasies.\footnote{Or on completely extraneous factors, such as
  budget constraints \citep{lange-2002}.} In this context work such as
that surveyed here can make a real contribution, by putting the
identification of senses on a much more objective footing, and also
allow for different sense granularities for different purposes by
adjusting model parameters \citep{erk-2010}.

On a more basic level, these questions are intimately related to some
of the basic theoretical and methodological conundrums of linguistics,
such as the nature of \emph{words}
\citep{aikhenvald-dixon-2002,haspelmath-2011}, of \emph{concepts}
\citep{murphy-2002,wilks-2009,riemer-2010} and their relation to word
meaning or \emph{word senses}
\citep{cruse-1986,kilgarriff-1998,kilgarriff-2004,hanks-2013}.

Relating the main kinds of lexical change which have been considered in computational linguistics to those discussed in historical linguistics (Section~\ref{sec:lingappr}), we note that there is no neat one-to-one correspondence. The study of \emph{semantic change} looms large in both fields and by and large focuses on the same kinds of phenomena, but in computational work, this is typically combined with a study of gain and loss of lexemes, since these phenomena are uncovered using the same computational methods. This could be said to constitute a consistent onomasiological focus, which however is not normally present in historical linguistics and consequently not given a label. In this survey, we refer to it as \emph{diachronic conceptual change}, i.e. change in the set of lexical meanings of a language. We propose the term diachronic conceptual change as a superordinate concept to semantic change. Diachronic conceptual change takes the view of all senses and word-sense allocations in the language as a whole. This includes a new word with a new sense (e.g., neologisms like \emph{internet} with a previously unknown sense) as well as an existing word with a new sense (\emph{gay} firstly receiving a `homosexual' sense, and later more or less losing its `cheerful' sense), because both of these add to the set of senses available in the language. Diachronic conceptual change also allows for changes to the senses themselves, the line between actual meaning change and usage change is blurry here. Examples include the \emph{telephone} that is a `device for conveying speech over a distance', but that is now also used for spread of communication, and increasingly as a `personal device used for photography, scheduling, texting, working', and so on. 

Further, the specific phenomena of \emph{lexical replacement} (including \emph{named entity change}) and its generalized version \emph{temporal analogy} have been subject to many computational linguistic studies.  Examples include \emph{Volgograd} that replaced \emph{Stalingrad} (named entity change), \emph{foolish} that replaced \emph{nice} for the `foolish' sense of the latter word (lexical replacement), and \emph{iPod} that can be seen as a temporal analog of a \emph{Walkman}.  
The change classes and their ordering as they are being studied from a computational perspective are shown in Table \ref{tab:changetypes-v2}.

\begin{table}[h]
    \centering
    \caption{Change types and their organization considered from a computational perspective. }\label{tab:changetypes-v2}
\begin{tabular}{ll}  
\toprule
\multicolumn{2}{c}{Lexical semantic change} \\
\midrule
Lexical change  &   Diachronic conceptual change\\
\cmidrule(r){1-1}
\cmidrule(r){2-2}
Lexical replacement &  Semantic change (new allocation between existing words and senses)\\
Named Entity change &  New word senses allocated to an existing word\\
Role changes &  New words with completely new word sense\\
Temporal analogues &  New word with a new but existing sense\\
& Changes to existing senses\\
\bottomrule
\end{tabular}
\end{table}

\section{Computational Modeling of Diachronic Semantics}\label{sec:wsch}
\subsection{Word-Level Change Detection}\label{sec:WLSC}
    
    The methods presented in this section aim to capture  diachronic conceptual change from a computational perspective and rely on different embedding techniques for representing words. While the papers surveyed in Section \ref{sec:NeurEmb}  feature neural embeddings, the papers surveyed in Section \ref{sec:emb} employ co-occurrence vectors in different ways. 
    All methods in this section represent all senses of a word using a single representation, that is, no sense discrimination or induction takes place. Within the sections, we have ordered the papers in diachronic order.   
   
  	\subsubsection{Co-occurrence-based methods}

   Most of the methods presented in this secion make use of co-occurrence information, and first build co-occurrence matrices. In some cases, the dimensions of the matrices are reduced using SVD. The majority use pointwise mutual information scores of different kinds (local, global or positive), rather than raw frequency scores for co-occurrence strength. Similarity is measured almost exclusivly using cosine similarity. \citet{rodda2016panta-journal} make use of second order similarity rather than work on first order similarity. \cite{KahmannNH17} use a rank series and compare differences in rank over time. Most different are the work of \citet{basilediachronic} that use random vectors to represent each word, and then context information, and \citet{Tang2013} that use contextual entropy and reduce dimensions on the fly rather than applying SVD as post-processing.
   
\paragraph{Context vectors}    \label{sec:emb}
    
    \citet{sagi2009semantic} presented work on finding the senses of words by using context vectors and found \textit{narrowing} and \textit{broadening} of senses over time by applying semantic density analysis. Each occurrence of a target word is mapped to its context vector, which follows the definition proposed by \citet{schutze-1998}.  A context is considered to be 15 words before and after each target word. 
    
    Two thousand words, the $50$th to the $2049$th most frequent word from the vocabulary are considered to be content-bearing terms $C$. Singular value decomposition is used to reduce the dimensionality by finding the most 100 important content bearing terms $C'$. \footnote{To capture as much variation as possible, the stopwords list is kept to a minimum and instead the 49 most frequent terms are ignored as terms not bearing any content. These can be considered to be stopwords in the specific vocabulary.}

For a specific target word $w$, each occurrence of the word in the corpus can be mapped to a context vector. The \textit{semantic density} of the word \textit{w} in a specific corpus is defined as the average cosine similarity of the context vectors. A high similarity can be seen as a dense set of vectors and corresponds to words with a single, highly restrictive meaning. A low similarity is seen as a sparse set of vectors and corresponds to a word that is highly polysemous and appears in many different contexts.  To reduce the computations, a Monte Carlo analysis was conducted to randomly choose $n$ vectors for pair-wise computation.
To measure change in word senses over time, context vectors are created for a target word in different corpora (from different time points) and the semantic density is measured for each corpus. If the density of a word increased over time then it is concluded that the meanings of the word have become less restricted due to a broadening of the sense or an added sense. Decreased density over time corresponds to a narrowing of the sense or lost senses.  
The authors used a set of 4 words in the evaluation on the Helsinki Corpus divided into four sub-corpora; \emph{do}, \emph{dog}, \emph{deer} and \emph{science}. The first two were shown to broaden their senses, while \emph{deer} was shown to narrow its sense. The word \emph{science} was shown to appear during the  period investigated and broaden its meaning shortly after being introduced. 

Unlike in the work by \citet{schutze-1998}, the context vectors were not clustered to give more insight into the different senses. Instead, a random set of context vectors were selected 
to represent the overall behavior of a word. This means that even though there can be indication of semantic change there are no clues as to what has changed. What appears as broadening can in fact be a stable sense and an added sense. In addition, the method requires very balanced corpora, because the addition of attributes such as genre will affect the density.

\paragraph{Pointwise mutual information}

Similar to the work described above, the work presented by \citet{Gulordava} builds on context vectors to identify semantic change over time. The authors used Google Books Ngram data, more specifically 2-grams (pairs of words) were chosen, so that the context of a word $w$ is the other word in the 2-gram. 

Two separate sub-collections were chosen, the first one corresponding to the years 1960--1964 (the 60s) and the second one corresponding to 1995--1999 (the 90s). The content bearing words were chosen as the same for both collections and each count corresponds to the local mutual information similarity score. Two context vectors corresponding to the word \textit{w} are compared by means of cosine similarity.

For a set of  10,000 randomly chosen mid-frequency words.  the similarity between the 60s and the 90s are computed. Out of these,  48.4\% had very high similarity scores, 50\% had mid-range similarity scores (between 0.8$-$0.2) and only 1.6\% had a  similarity score lower than 0.2. The assumption was that words with low similarity scores are likely to have undergone a semantic change, an assumption that was tested by manually evaluating a sample of 100 words over all similarities. Five evaluators judged each of the words on a 4-point scale (from \textit{no change} to \textit{significant change}) based on their intuitions. The average value of these judgments were then used for each word and compared using the Pearson correlation measure. 
The results show that distributional similarity correlates the most (0.445) with words that were more frequent in the 90s, while the frequency method correlates the most (0.310) with words that were more frequent in the 60s. It is important to note that the evaluation measured the ability to detect not only change, but also to distinguish the degree of change. For better comparison with other surveyed methods, it would be useful to see how this method performs for the 100 most changed words, and as a comparison, to the 100 least changed words. 

\citet{rodda2016panta,rodda2016panta-journal} present a method that investigates the change in ancient Greek from the pre-Christian era (7th c. BCE to the 1st c. BCE) to the centuries after Christianity (1st c. CE to the 5th c. CE). The study has its starting point in the idea that the rise of Christianity had a widespread effect on the Hellenic world. The method relies on second-order similarities on the basis of positive pointwise mutual information scores. A set of 50 words were investigated in detail by comparing their nearest neighbors between the two periods. 

The corpus is based on the \emph{Thesaurus Linguae Graecae} collection of Ancient Greek literary texts excluding \eg private letters.
The documents were lemmatized, stopwords were removed, and lemmas that occurred less than 100 times were filtered out. 
Among the remaining words, there were roughly 4,000 lemmas (we call these $V$) that co-occurred in both sub-corpora and were used for the second-order similarity analysis. 

The method creates a first order co-occurrence matrix using positive pointwise mutual information scores for each of the two sub-corpora. After Singular value decomposition, the dimensions were reduced to 300, that could possibly be disjoint. Next two second-order similarity matrices, $S_1$ and $S_2$, were calculated using cosine similarity. Finally, the semantic similarity of a word  was calculated as the Pearson correlation between the corresponding vectors in $S_1$ and $S_2$. 

The analysis was performed by manually evaluating the 50 lemmas with the lowest correlation score between the two sub-corpora. Each word was analyzed by comparing the 10 nearest neighbors from $S_1$ and $S_2$. The authors noted that there were two prominent groups with changed meanings among these 50; Christian terms (\eg \emph{la\'os} that goes from `people' to `Christians') and technical terms (\eg \emph{hyp\'otasis} that goes from `foundation' to `substance').

A recent paper that relies on co-occurrence information was presented by \citet{KahmannNH17}. The authors propose using \emph{context volatility} based on the significance values of a word's co-occurrence terms and their corresponding ranks over time. The method starts by calculating a co-occurrence matrix and then transforming this matrix to a significance matrix, where \eg the pointwise mutual information score of the two words was used instead of their co-occurrence frequency. For every set of co-occurring terms \textit{$w_i$} and \textit{$w_j$}, the context volatility measure, proposed by \citet{Heyer2016} for detecting changes in topics, calculates their rank series at every time slice \textit{t}. The average coefficient of variation for every time series of a fixed word $w_i$ with all words $w_j$ is the volatility of $w_i$. \citet{KahmannNH17} propose an extension to the context volatility methods, called the MinMax algorithm, which takes into consideration gaps (i.e., missing co-occurrence values in the time series of $w_i$ and $w_j$, e.g., due to sparse data). The MinMax algorithm measures the mean distance between the ranks of the time slices.  
 The authors present experiments on an artificially created dataset to show three change types; (1) a change in the probability of a context term, (2) novel contexts, and (3) disappearing contexts in which the MinMax algorithm shows performance that is most resilient against differences in term frequency and the different types of modeled change. 
 
 The authors also evaluated a set of four terms in a British newspaper corpus collected from the Guardian between January and November 2016, around the terms \emph{brexit} and \emph{referendum}. The MinMax algorithm showed volatility for the terms \emph{brexit}, \emph{cameron}, \emph{may} and \emph{farage}, the latter three corresponding to David Cameron, Theresa May, and Nigel Farage who all played an important role in the time leading up to the Brexit referendum. The method also showed correlation between the frequency of the terms \emph{brexit} and \emph{cameron} and their volatility.   
 
 The authors claim two main advantages of the method, first, it can overcome the sparsity problems of diachronic corpora by not relying on (neural) embeddings that require large amounts of data, and second, it does not need to start with a fixed set of terms but rather detects the most changed terms. The latter does not seem to hold, since most published work make use of the "top" output of their methods to find the words most likely to have changed (see Table \ref{tab:compOfWSCmethods}).

\paragraph{Temporal Random Indexing} 

\citet{basilediachronic} presented one of few studies of the semantic change problem in a language other than English. They focused on Italian and released a set of 40 words with their corresponding shifts in meaning. 
They made use of a word embedding method called Temporal Random Indexing that builds on the authors' previous work \cite{Basile-2014}. They made use of the Italian portion of the Google Books Ngrams corpus and considered the whole corpus to create the vocabulary \textit{V}. Each term in \textit{V} gets a randomly assigned vector with two non-zero elements $\in \{-1,0,1\}$ and the assignments of all vectors corresponding to \textit{V} are near-orthogonal. 
Then the corpus is split into sub-corpora where each one corresponds to a 10-year period between 1850--2012. 
The vocabulary in each sub-corpus is then modeled as the sum of all the random vectors assigned to each context word, 
the sums are normalized 
to downgrade the importance of the most frequent words.

The method can be repeated iteratively to update the representation of a word with more context information, but the authors found that the iterative vectors (both one and two iterations) gave worse results than those that used only the random vectors. 
The log frequency was used as a baseline for each word. The authors then used the change point-method proposed by \citet{kulkarni2015statistically} with two versions for detecting change, the pointwise change between two time adjacent vectors (\textit{point}) ($t_{i-1}, t_i$) and the cumulative change (\textit{cum}) between the sum of all vectors up to $t_i$ and the vector for $t_{i+1}$. \footnote{Note that this is different from \citet{kulkarni2015statistically} who 
compare to $t_0$ for all time points $t_i$} 

A list of words that have changed was created, together with all the change points corresponding to each word, by pooling the top 50 results of each method and manually checking the pooled results against two dictionaries. The resulting dataset consisted of 40 words.  
An evaluation was performed manually on the 40 words. Given the set of change points returned by each method, the evaluation checked how many correct change points were detected among the top 10, top 100 and all of the returned change points. A change point is considered correct if it is found at the same time, or after the expected change point. 

At the top 10 and top 100, the accuracy of the random indexing method with \textit{point} performed as well as the baseline, and both outperformed the other methods. When considering all change points, the random indexing with \textit{point} performed best, followed by random indexing with \textit{cum}. 
The authors presented a time-aware evaluation as well and evaluated in which time delay the change points were found. The temporal indexing with \textit{point} that got the best top 10 and overall scores had a time delay of, on average, 38 years with a standard deviation of 35. The best results were obtained by the random indexing with one and two iterations and the \textit{cum} that, on average, had a delay of 17$\pm$15 and 19$\pm$20 respectively. These methods, however, had only an accuracy of 12--16\% on the detected change points.

\paragraph{Entropy}
\citet{Tang2013} presented a framework that relies on time series modeling on the changes in a word's contextual entropy. The method itself is not sense-differentiated, but from the time series, the authors showed patterns that differentiated broadening and narrowing, novel senses, as well as metaphorical and metonymic change.

For each period, a word was modeled as a distribution over its strongest noun associations. 
We can view this procedure as analog to first calculating a co-occurrence matrix and then performing dimensionality reduction, but here the dimension is reduced directly by associating \textit{w} only to one noun from each context.  The authors claimed that this helps represent different sense, as nouns have a high differentiating value.
A \textit{word status} for \textit{w} at time \textit{t} is then the probability of these contextual nouns.
To create a time series, the feature vectors are represented by their entropy. The authors model linguistic change as an S-shaped curve \citep{kroch_1989} and apply curve fitting on the time series of the word status entropy 
 to detect patterns for different kinds of change. 

The Chinese newspaper \emph{People's Daily}, spanning 1946--2004, a total of 59 years with approximately 11 million tokens. They chose 33 words from Chinese new word dictionaries to represent changed words and another 12 words, not in any new word dictionary, as examples of stable words. The authors first noted that the S-variable could be used to determine if a word had changed its semantics. 
They also noted that not being found in the new word dictionaries did not preclude a word from having changed semantics, and upon manual inspection, found that 5 out of 9 stable words had also changed their meaning. 13 out of 19 words that had experiences broadening and narrowing could be found, as well as 3 out of 5 words that experienced metaphorical change.

This experiment shows that the entropy time series of a word's feature vector can be used to identify different kinds of change. However, the values used for the classification are observations from the training data (all words that are classified have also contributed to the finding of the thresholds). The experiment does not show the discriminating power of the 
variables on previously unseen data. 

In a follow up work, \citet{Tang2016} attempted to cluster the contexts to find senses, and to classify the senses into different change types.\footnote{This approach is considered sense-differentiated but we place it here since the description of the main algorithm is here.}  In this experiment the dataset was larger and consisted of 197 words that had experienced change. This change vs. stable experiment was conducted using a training and a testset where 28 stable terms and 28 changed terms were chosen randomly. By separating the dataset into separate training and testsets, the weakness of the previous paper is overcome. The experiment was repeated 100 times and the average precision of the experiment was 82.3\%, using Support vector machine (SVM) for the classification.  

In a second step, the contexts for each word were clustered using the DBSCAN algorithm. The resulting clusters were considered synsets. 
The resulting set of synsets is large; for example, for the word \textit{tou ming} there were 490 clusters, so there is a need for reduction. The authors proposed using the diachronic span and density of each cluster to determine whether the cluster is sufficiently strong for continued analysis, or should be discarded. This procedure resembles word sense induction, and thus is considered as sense-aware even if the relation between the resulting synsets and the word senses were not directly clarified or evaluated. It is also unclear how many synsets remain on average after the reduction phase. 

The individual senses were classified on the basis of the sense status using the same procedure used for classifying change vs. stable words. The classification was done for three change types at the same time (novel sense, metaphorical change, and metonymical change). They used 14 randomly chosen training examples for each type. The remaining 100 or so elements were used for testing. The experiment was repeated 100 times and the average precision was 42\%. The results were not reported for each class individually and thus we do not know if some classes performed better than the others.  The authors concluded that while it is possible to distinguish the different classes for each synset, the variables of the S-shaped curve were not sufficient for accurate classification. 
One important weakness  is that the model only allows for one change event per word or sense. It is, however, possible that more than one change event occurs for each sense. In addition, the sense induction procedure was not evaluated properly; a different induction method (i.e., a different grouping of nouns into synsets) might provide better results.

 \subsubsection{Neural Embeddings} \label{sec:NeurEmb} 
     In the last few years, the largest body of work has been done using (neural) word embeddings of different kind. 
      With a few exceptions, embeddings are individually trained  on different time-sliced corpora and compared over time. This means that each representation for a word at different points lives in a different space. All different embeddings for a word must first be projected onto the same space before comparison. Three different methods have been used for projection.
      First, vectors are trained for the first time period $t_1$ independently of any other information. The follow up vectors for time $t_i \forall i > 1$ are initialized with the vectors for time $t_{i-1}$. What happens in the case of words that are present in $t_i$ but not $t_{i-1}$ is generally not specified, the same initialization can be assumed as at time $t_1$ (see \eg \citet{TempAnal-NeuralLangModel} for more details).
      The second method projects words to a specified time period, typically the last one, using a linear mapping (see \eg \citet{kulkarni2015statistically, DiachronicWordEmb} for more details and examples). 
      Finally, the third method avoids mapping of vectors by comparing second order similarity vectors \citep[see][]{Eger-SemChange}.  All of the papers in this section consider time series data and make use of different methods to detect changes compared to the average, first or last time period.

\paragraph{Initializing using previous time period}
 \citet{TempAnal-NeuralLangModel} were the first to use neural embeddings to capture a word's meaning for semantic change detection.  They used the Skip-Gram model \cite{Mikolov2013Efficient} trained on the Google Books Ngrams (5-gram) English fiction corpus. They created a neural language model for each year (with 200 dimensions), with the vectors being initialized by the vectors from the previous year. Years 1850--1899 were used as an initialization period and the focus for the study was 1900--2009. Vectors were compared over time using their cosine similarity. The 10 least similar terms (those believed to have changed their meaning the most) and the 10 most similar terms (stable terms) were inspected. 
The three closest neighboring words from 1900 and 2009 were used for verification of change.

Two words were investigated in more detail with respect to the time series of their cosine similarities; the difference in cosine similarity between the year \textit{y} and 1900 was plotted against a time axis. This was compared to the average cosine similarity of all words as a baseline. It was clear that \textit{cell} and \textit{gay} deviated significantly from the average plot while the two stable terms \textit{by} and \textit{then} were more stable than the average. The comparison to the average of all words is an alternative method to comparing to negative words. This controls for the fact that not all words behave the same way as the changing ones and thus confirms the correct hypothesis. An alternative method  is to compare, not to the overall average, but to the average of words in the same frequency span as the word under investigation (like in \citet{jatowt2018}). Comparing to other words in the same frequency span is important as there is evidence that very frequent words behave differently from very infrequent terms \cite{DiachronicWordEmb,pagel2007frequency,lieberman2007quantifying}. 

The two changing terms were also investigated in a pairwise manner with respect to four terms that represented their different senses. \textit{Cell} was compared to \textit{closet} and \textit{dungeon} on the one hand and \textit{phone} and \textit{cordless} on the other. The term \textit{gay} was compared to \textit{cheerful} and \textit{pleasant} as well as \textit{lesbian} and \textit{bisexual}. 
 In addition, the authors further grounded their results by investigating ngrams that contained the evaluated word from 1900 and 2009. We note that this, backwards referral to the original texts that contribute to a statistical hypothesis, is an extremely important step that is often overlooked by others. 
     
     The authors concluded that a word that has lost in popularity over time, and hence is not frequently used in succeeding time spans, will not update its vector and, therefore, change cannot be detected. They suggest combining embedding signals with frequency signals to detect such cases. 
      No explicit evaluation with respect to outside ground truth was made, nor were the words marked for being correct or incorrect. Finally, the authors concluded that their approach does not distinguish change types such as broadening, narrowing or pejoration, and amelioration, yet they suggest it would be interesting to analyze and characterize these.

\paragraph{Change point detection}
\citet{kulkarni2015statistically} presented an investigation of different statistical properties and their capacity to reflect statistically significant semantic change. Two questions were asked; how statistically significant is the shift in usage of a word over time? and at what point did the shift take place? Two things seem to be implicit in these questions. First, a shift in the dominant sense of a word (\eg one existing, dominant sense handing over to another existing sense) was also considered a semantic shift. And secondly, a word has only one semantic shift. The authors noted that while many semantic changes occur gradually, there is a time when one sense overtakes the others and they considered this to be the change point, on lines of explanatory power of a  sense (see section \ref{sec:evaltechWSE}). 

For each word in a vocabulary, a time series was constructed over the entire time period. Each point $t$ in the time series results in statistical information derived from the word's frequency, its part-of-speech distribution, or its semantic vector, corresponding to the usage in the sub-corpus derived at $t$, namely $C_t$. 
Once all time series have been constructed, the authors proposed a change point detection method.  
 The  time point $j$ that produces the most significant change point (with a score above a user-defined threshold ) was considered the final change point. 

Three datasets were used, The Google Books Ngram Corpus (5-grams from 1900--2005, 5-year periods), an Amazon Movie Review corpus (1997--2012, yearly periods), and a Twitter corpus (2001--2013, monthly periods). 
A set of words were investigated in more detail and it was found that the distributional method performed better for some (a set of 11 words like \textit{recording, gay, tape, bitch, honey}) while the syntactic method performed better for others (\eg
\textit{windows, bush, apple, click}). For the three first (\textit{windows, bush, apple}), the distributional method could detect significant changes, even though the most common context words should have been significantly different in the ages of computers and presidents. The word \textit{bush} provided a good reason for allowing more than one change point given that the US had two presidents with that name, but in rather different contexts. 

A synthetic evaluation was presented, in which 20 duplicate copies of Wikipedia were used 
and the contexts of the words were changed artificially proportionally to a probability $p_{r}$. 
The larger the proportion of $p_{r}$, the better did both the distributional and the frequency method perform, with the distributional method outperforming the frequency method for all values of $p_{r}$.
When the target and the replacement words were no longer required to belong to the same part of speech, the distributional method was outperformed by the syntactic method for low values of $p_{r}$ but for values $ \geq 0.4$ the distributional method outperformed the syntactic method. 

The second evaluation was performed on a reference set of 20 words, compiled from other papers. %
Out of 200 words evaluated per method, 40\% of the words from the reference set were found for the distributional method and 15\% for the syntactic method. This was to some extent, an experiment to capture recall of known changes.  

Finally, in the human evaluation, the top 20 words from each method were evaluated by three annotators. For the frequency method, 13.33\% of the 20 words were considered to have changed their meaning, for the syntactic method the corresponding number was 21.66\%. For the distributional method the rate was 53.33\%, meaning that among the top 20 words outputted by the distributional method, more than half were deemed to be words that had undergone semantic change. This was however not tied to the change point; a word was only judged to belong to either class \textit{change}/\textit{no change}. An interesting question arises when the time series of the syntactic and distributional methods were created. For both, the data at time $t_i$ were compared to $t_0; \forall i$, where $t_0$ corresponds to the earliest possible time point and might have low quality due to sparse data and a high error rate. Would the method perform better if the information at $t_i$ were compared to $t_N$ where $N$ was the last time point, to $t_{i-1}$, to an average of all time points, or to a joint modeling of all information at once?

\paragraph{PPMI-based compared to SGNS}      
\citet{DiachronicWordEmb} presented an evaluation of different embedding techniques for detecting semantic changes, both a priori known pairs of change and those detected by the different methods. 
They evaluated six different datasets, covering four languages and two centuries of data. 

The first embedding method was based on the positive pointwise mutual information score PPMI. The second was a Singular value decomposition (SVD) reduction of the PPMI matrix, often referred to as SVD$ _{PPMI}$ in other work, and the third embedding method was the Skip-Gram with negative sampling (SGNS) \cite{Mikolov2013-Distributed}. For the two latter, embedding dimension was 300 with a window size of 4 (on each side). 
The SVD and SGNS embeddings were aligned over time using the orthogonal Procrustes. 

Four different tasks were evaluated; synchronic accuracy, detection of known pairs of change on both COHA and ENGALL, Google Books Ngram all genres, and discovery of new words that have changed on ENG fiction.  
The synchronic task was not relevant for change detection, but the SGNS performed worst out of the three measures on ENGALL. 
The pair-wise task considers the cosine similarity of a pair of words at each time slice, and correlates the value against time using Spearman correlation.

For the detection of known pairs, a set of nine terms were compared with respect to a validation word. As an example, the term \textit{nice} should move closer to \textit{pleasant} and \textit{lovely} and away from \textit{refined} and \textit{dainty}, resulting in four pairs. A total of 28 pairs were evaluated on both ENGALL and COHA.
All three measures were able to detect all changes for COHA and ENGALL except for PPMI that captured 96.7\% of all changes on ENGALL. The differentiating factor became the number of significant changes that were found for the different measures; PPMI scores lowest for both datasets with 84--88\% significant changes detected, SVD performed best with 90--96.0\% (ENGALL and COHA) while SGNS performed best on ENGALL with 93.8\% and lowest on COHA with 72.0\%. This was likely a result of the size of COHA that was insufficient for SGNS.

On the detection task, the top 10 words were evaluated for all three methods on ENGfiction and Google Books Ngram fiction, the authors note that the top 10 words on ENGALL are dominated by scientific terms due to sampling biases in the corpus. The 30 words were evaluated by evaluators and classified as correct (i.e., true semantic change, borderline and incorrect). For SGNS 70\%  (that is, 7 out of 10 words) were correct, for SVD 40\%  were correct and for PPMI only 10\%  were correct. However, for PPMI, the remaining nine words were deemed as borderline while SVD had two borderline cases and SGNS has only one. So, considering the inverse account, SGNS had 20\% incorrect, SVD had 40\% incorrect and PPMI had zero incorrect results among their top 10 ranked words. 
It would be interesting to know the results on COHA, and if SGNS performs worse than SVD on this smaller dataset.

\paragraph{Second-order similarity}
\citet{Eger-SemChange} presented a method that relies on second-order similarity of Word2Vec-vectors. The primary aim was to investigate whether semantic change is linear over the vocabulary as a whole, but the method can also be used to detect words that deviate from the pattern of general change. 

The COHA corpus was used as a basis with two partitions -- 1810--2000 and 1900--2000, respectively -- divided into decades. For each partition, only nouns, adjectives, and verbs were kept that occurred more than 100 times. The experiments were repeated on a German newspaper corpus, \emph{S{\"u}ddeutsche Zeitung}, for a yearly span in the period 1994--2003, and a Latin corpus covering the years 300--1300 subdivided into 100-year periods. 

The method starts with a common vocabulary $V$ that is the intersection of all words in all time sliced sub-corpora. For each word in $w_i\in V$, an embedding was created $\tilde{\textbf{w}}_i(t)$ from the sub-corpus corresponding to time $t$. To avoid mapping or transforming vectors across time, the (second-order similarity) embedding of a word $\textbf{w}_i(t)$ was based on the cosine similarity between $\tilde{\textbf{w}}_i(t)$ and all words $\tilde{\textbf{w}}_j(t), j=1\ldots|V|$. 

The hypothesis is that meaning change can be modeled by a linear model, that says the meaning of a word $w_i$ at time $t$ can be modeled as a linear combination of its neighbors from previous time points. 
To test this hypothesis, a time-indexed self-similarity graph was used. A self-similarity graph counts the similarity between the embeddings of two time slices for the same word, sim($\textbf{w}(s), \textbf{w}(t)$) for any two $s,t$. The self-similarity was then averaged for all time gaps $t_0=|t-s|$. 
Across all words and all datasets, self-similarity shows a linear decay with time distance, and the linear model with a negative coefficient, fits with $R^2$ values of well over 90\%. To measure change over time, a ratio of 
the maximal and minimal weight of a link in the self-similarity graph of $w$ was used. For the 1900--2000 COHA dataset, the bottom words were \textit{one, write, have, who, come, only, know, hat, fact}, and among the top ones  were words like \textit{bush, web, implement, gay, program}, showing the potential for discriminating between stable and changing words. 

The authors investigated the possibility of finding words that have negative relationships, that is, those that move apart. Four pairs were shown as examples, summit $\leftrightarrow$ foot, boy  $\leftrightarrow$ woman, vow  $\leftrightarrow$ belief and negro  $\leftrightarrow$ black. The paper provides justification only for the last pair.

\paragraph{Concept change}
\citet{recchia2016tracing} claim that semantic change and concept change are different in that a concept can connect more than one sense. For example, the main sense of the concept of \emph{broadcast} changed from `spreading seeds' to `spreading information' but the concept does not lose the sense of spreading seeds. Instead, the second sense becomes more prominent. Their use of ``concept'' is idiosyncratic, and corresponds roughly to the lexical-semantic relation superordinate or hyperonymy or the lexicographic notion of main or prototypical sense \citep{lew-2013}.
To be able to detect changes in concepts, the authors proposed creating a fully connected graph around a seed word. The graph is created in the following way; first, all words are represented by their HistWord vectors \cite{DiachronicWordEmb}. Second, two words are linked, if the cosine similarity of their corresponding vectors is above a threshold, considered as the minimal edge weight. 

Once the fully connected graph (size 9) has been created for the first time slot, a word is exchanged in a subsequent time slot if and, only if, it increases the minimum edge weight, and the word that leads to the highest increase is chosen. At each time period (decade), at most one word from the fully connected graph is exchanged. 

On average, 33\% of the fully connected graphs shared at least one word in common between the earliest (1800--1810) and the last (1990--2000) time period. In 69\% of all cases, the seed word was no longer present in the last time period. The authors noted that the graphs did not typically drift too far afield, showing that the concepts are flexible but ``reasonably resistant'' to drift. 

There was no systematic evaluation to confirm this claim. The authors noted that out of 500 randomly chosen words, 
only 212 had a fully connected graph where the minimum edge weight exceeded 0.2. The method should thus be applicable to roughly 42\% (= 212/500) of the words. Still, the idea of tracking broader sense clusters rather than individual senses is highly appealing.

\subsubsection{Dynamic word embeddings}
Three different methods exist for creating dynamic word embeddings. Common to all of them is that they share some data across all time periods and that the resulting embeddings originate in the same space for all time periods. This reduces the need to align the vectors trained on separate time slices. However, each method uses different embedding techniques. It shows that, regardless of method for creating individual embeddings, sharing data across time is highly beneficial and can help reduce the requirments on large datasets (which we rarely have available for historical, textual corpora). 

\paragraph{Dynamic probabilistic Skip-Gram}
The paper by \citet{bamler17} was the first of three to propose using dynamic word embeddings trained jointly over all times periods. The advantage of the method is two-fold. First, there is no need to align embedding spaces which can introduce noise, and second, the model utilizes information from all time periods to produce better embeddings and reduce the data requirements. 

The authors proposed a Bayesian version of the Skip-Gram model \cite{Barkan-probSKipgram} with a latent time series as prior. Their method is most like that of \citet{TempAnal-NeuralLangModel}, but information is shared across all (or all previous) time points. The priors are learned using two approximate inference algorithms, either as a \textit{filtering}, where only past information is used (for time $t_i$ all information from $t_0$ to $t_{i-1}$ is used), or as a \textit{smoothing}, where information about all documents (regardless of time) is used. The resulting dynamic word embeddings can fit to data as long as the whole set of documents is large enough, even if the amount of data in one individual time point is small. 

The authors compare their methods, dynamic Skip-Gram with filtering (DSG-f) and smoothing (DSG-s) with the non-Bayesian Skip-Gram model with the transformations proposed by \citet{DiachronicWordEmb} (SGI) and the pre-initialization proposed by \citet{TempAnal-NeuralLangModel} (SGP). Three different datasets were used, Google Books corpus with 5-grams (context window size of 4), the State of the Union addresses of U.S. presidents (spanning 230 years), and a Twitter corpus spanning 21 random dates from 2010 to 2016. For the first corpus, the dimension size was 200 and for the two smaller corpora, the dimension size was 100. 

The quantitative experiments aimed to investigate the smoothness of the embeddings between different, adjacent time periods. 
The first method visualized embedding clouds for the Google Books Ngram dataset for four consecutive years. 
The second method shows the displacement, measured as a Euclidean distance, between word vectors in  1998 and the next 10 years.
Finally, the third method investigated the generalization to unseen data and showed that DSG-f and DSG-s outperformed SGI and SGP for the Twitter and SoU corpora. 
The experiments showed that joint training over all time periods is beneficial when training vectors for individual time periods, in the sense that the vectors do not move too radically from one year to another. 

The second set of experiments aimed at showing the capability of detecting semantic change. First, the top 10 changing words according to the DSG-f method were visualized by means of their five most similar words from the earliest and the last time points. The reader is left to judge for themselves if all words are indeed correct. 
Second, three changing words are investigated with respect to two opposite context words (in pairs) for each corpus. As an example, \textit{computer} was compared to \textit{accurate} and \textit{machine} for the Google Books dataset. Again, the DSG-f and DSG-s outperformed SGI and SGP for the two smaller datasets, where the latter two methods had difficulty fitting a vector space to small amounts of data. For Google Books, the dynamic embeddings performed better in the sense that they were smoother. \footnote{There are no precision values in the paper; the interpretation of the change results is left to the reader.}
 
\paragraph{Dynamic PPMI embeddings}
\citet{Yao-2018} presented a second approach, with a different take on the word embeddings. Their embedding method relies on a positive pointwise mutual information matrix (PPMI) for each time period, which is learned using a joint optimization problem. In other words, embeddings for each time period were not first learned, then aligned, but rather learned while aligning. 

The authors proposed these dynamic embeddings for both the semantic change problem and the diachronic word replacement problem. They investigated both problems using qualitative and quantitative evaluation. The authors crawled roughly 100k articles from the New York Times, published 1990--2016, together with metadata such as section labels. Words that occurred fewer than 200 times across all time slices, as well as stop words were removed. Using a vocabulary of almost 21k words, and a window size of 5, the embeddings were created.

Four words were evaluated manually. 
The first two clearly illustrate the difficulty with modeling a word's meaning using a single representation: \emph{apple} has nothing to do with fruit from 2005 and onward and, since 1998, \emph{amazon} is not a river in South America.\footnote{In this dataset, the confusion of "apple" as a fruit and "Apple" as a company could be a consequence of  the case normalization preprocessing step. The case of "Amazon" is different, since both the jungle and the company are proper nouns. It might also be a consequence of a change in the dominant sense of the words, from fruit and a jungle to a company, and the representation method that might have difficulty capturing both at once.  } 

For the automatic evaluation, the authors created a ground truth dataset using the section category of the 11 most discriminative categories.
The authors then clustered the embeddings of each word using a spherical k-means (using cosine similarity between the embeddings as a similarity measure) and k = 10, 15, and 20. The clusters were valuated using two metrics, normalized mutual information (NMI) between the set of labels and the set of clusters, and F-measure. The F-measure considers any two word -- time pairs; if they are clustered together and have the same section label, then the decision is correct, and otherwise the decision is incorrect. 

The comparison is done against three baselines,  Static-Word2Vec (Sw2v) \cite{Mikolov2013-Distributed}, Transformed-Word2Vec (Tw2v) \cite{kulkarni2015statistically} and Aligned-Word2Vec \cite{DiachronicWordEmb} (Aw2v). Both NMI and F-measures showed that the dynamic embeddings were better than the baselines, and while Sw2V and Aw2v followed closely, Tw2v showed a larger drop in performance. The authors suggest that this happened because local alignment around a small set of stable words was insufficient. While this seems reasonable, it does not explain why the Sw2v method (without alignment) performs better than the Aw2v method for all values of $k$ for the NMI measure and was worse only for $k=10$ for the F-measure. 

\citet{Yao-2018} do not reference to the work of \citet{bamler17}, and despite the different publication years, the work of \citet{Yao-2018} was submitted  before the work of \citet{bamler17} was published. Nonetheless, there is much overlap in the idea of jointly learning and aligning temporal vectors to produce smoother vector series for individual words. Since information regarding most of the vocabulary is shared across time slices, the dynamic PPMI embedding method is considered robust against data sparsity, however, the authors did not mention any size requirements. 

\paragraph{Dynamic exponential family embeddings}
A third method for creating dynamic embeddings was presented by \citet{RudolphB18-dynamicEmbforLangEvo}. This method makes use of exponential family embeddings as a basis for the embeddings, as well as a latent variable with a Gaussian random walk drift. 
The key is to share the context vectors across all time points, but the embedding vectors only within a time slice. The results were compared to the results presented by \citet{DiachronicWordEmb} and the exponential family embedding (the static version).  

The authors used three datasets, machine learning papers from ArXiv (2007--2015), computer science abstracts from ACM (1951--2014), and U.S. Senate speeches (1858--2009). The 25,000 most common words were used, and the others were removed. In addition, words were removed with a probability 
 proportional to their frequency to downsample frequent terms and to speed up training. The embeddings had a dimension of 100. 

As with \citet{bamler17}, the dynamic embeddings performed better on unseen data. 
In a qualitative setting, a set of six example words were used to illustrate semantic drift, where the meaning of a word can change; its dominant sense can change; or its related subject matters can change. There was no explicit differentiation between the change types. Instead, 
the absolute drift was computed as the Euclidean distance between the first and the last time points. Note that if the curve of changes in the embeddings behaves like a sine curve, there can be little difference between the first and the last change point, and the word can still experience substantial semantic drift in between.
The authors presented the 16 words with the highest drift values for the Senate speeches, and discussed a few of them in detail. They did however  not present their view of these 16 words, or if any were considered incorrect. 

A change point analysis was presented, and contrary to \citet{kulkarni2015statistically}, the authors did not make an assumption of a single change point, but no change point evaluation was presented. They calculated the time point where a word $w$ changed the most, by normalizing with the average change of the others words $v$ in the same time point. 

A novelty of this paper is the investigation into the distribution of those words that changed the most in a given year. It does give some account of where interesting things happen to the language as a whole, and the authors recognize that the largest change occurred at the end of World War 2 (1946--1947), for the Senate speeches. Another interesting spike occurred in 2008--2009 and what seems as the 1850s but these were not discussed further. 
The authors conclude by noting that the closest neighboring words over time show the semantic drift of words and can be a helpful tool to discover concept changes.

\subsubsection{Laws of sense change}
Several authors have investigated general laws of sense change from large corpora. Here we summarize these laws. 

\citet{Xu15-twolawsofsemChange} evaluated two laws against each other, with respect to synonyms and antonyms. Using normalized co-occurrence vectors and the Jensen-Shannon divergence, \citet{Xu15-twolawsofsemChange} investigated the degree of change for a given word measured as the difference in overlap between the nearest 100 neighbors from the first and the last year of the Google Books Ngrams corpus. Using a set of synonyms and antonyms and a set of control pairs, the authors showed that, on average, the control pairs moved further apart than the synonyms and antonyms. They call this the \textit{law of parallel change}, words that are semantically linked, like synonyms or antonyms, experience similar change over time and thus stay closer together than randomly chosen words.  

\citet{Dubossarsky-2015} investigated the relation between a word's role in its semantic neighborhood and the degree of meaning change. Words are represented using their Word2Vec vectors trained on a yearly sub-corpus and similarity is measured using cosine similarity. Each yearly semantic space is clustered using k-means clustering (this can be seen as word sense induction but without the possibility for a word to participate in multiple clusters). A word's  \textit{prototypicality} (centrality) is measured as its distance to its cluster centroid (either a mathematical centroid, or the word closest to the centroid). Change is measured as the difference in cosine similarity for a word's vector in adjacent years, where the vector of the previous year is used as an initialization for the next, as in the work of \citet{TempAnal-NeuralLangModel}.  The correlation between a word's centrality and its change compared to the next decade is measured. The 7,000 most frequent words in 2nd version of the Google Books Ngrams English fiction corpus were investigated.

The authors showed that there is a correlation between a word's distance from the centroid and the degree of meaning change in the following decade. The correlation is higher for the mathematically derived centroid, compared to the word closest to the centroid. This indicates that the abstract notion of a concept might not necessarily be present as a word in the lexicon. Also the number of clusters play a role. In this study, the optimal number of clusters was 3,500, but this should reasonably change with the size of the lexicon. The trend was shown for a large set of words (7,000) over a century of data. This is the \emph{law of prototypicality}. 

\citet{DiachronicWordEmb} suggested two laws of change, the \emph{law of conformity}, which states that frequently used words change at slower rates, and the \emph{law of innovation}, which  states that polysemous words change at faster rates. Polysemy is captured by the local clustering coefficient for a word in the PPMI matrix, which captures how many of a word's neighbors are also connected as a proxy for the number of different contexts that a word appears in. 

At the same conference as \citet{DiachronicWordEmb}, \citet{Eger-SemChange} presented the \textit{law of linear semantic decay} which states that semantic self-similarity decays linearly over time. They also presented the law of differentiation, which shows that word pairs that move apart in semantic space can be found using the linear decay coefficient.

 \subsubsection{Related technologies}

 \citet{Mihalcea_wordepochdis} investigated the effect of word usage change and formulated an inverse problem to identify the epoch to which a target word occurrence belongs. This was done as a classification task (word sense disambiguation) using three epochs. 
 The word sense disambiguation algorithm chose a set of local and global features; \eg part of speech (of the word and the surrounding words), the first verb/noun before and after, and topical features determined from a global context. Overall, 
 showed an improvement of 18.5\% over the most-common sense 
 baseline for all test words, and the largest improvement was found for nouns. The authors observed that words with a high frequency in one epoch that had experienced semantic (or usage) change improved greatly over the baseline. The authors also noted a difference in results for polysemous words compared to monosemous words and took this as an indication that semantic change is reflected in the context (claiming that monosemous words do not experience semantic change over time).

 \citet{JapnLoanwords} targeted a slightly different but related task to identify the difference in meaning between Japanese loanwords and their English counterparts. This is done by creating embeddings (Word2Vec, Skip-Gram  with negative sampling and dimension sizes between 300--600) for loan words in two different corpora, namely a snapshot of the English and Japanese Wikipedia. They projected the Japanese embeddings onto the same space as the English embeddings. If the embedding for a Japanese loanword and its corresponding original English word were far apart (measured by cosine similarity), then they were likely different in meaning. 
 The authors recognized that semantic change in this context could mean that the Japanese loanword only adopted a single sense from a word's senses.  Hence, the embedding created for the English corpus (where all senses are present) was different from that created for the Japanese corpus, where only one or a few senses was present.

	A method to go beyond pure vector changes and look at the surrounding words is proposed by  \citet{vanAggelen}. They linked embeddings to WordNet to allow quantitative exploration of language change, for example, to which degree the words in one part-of-speech change over time. \citet{hamilton-cultShift} attempted to show that semantic change can be differentiated as reflecting (pure) lexical or cultural change. However, \citet{vanAggelen} could not replicate the experiment, which means further investigation, possibly into additional languages, is needed. 
\citet{Chiru-2014} made use of the visual trends of words belonging to three classes, neologisms, archaisms and common words. For words in each class, which were chosen from a predefined list, a search was made on Google Ngram viewer and the visual trend for the class was calculated using principal component analysis. For a new word, the frequency graph was transformed and compared to the trend of each class. The word was assigned to the class in which the trend fit best. Out of 414 neologisms (which also had a sufficiently high frequency in Google Books), 334 (81\%) were correctly classified as neologism, 64 (15\%) as common words and 16 (4\%) as archaisms. 
In addition, \citet{tjongkimsang} made use of frequencies to detect neologisms and archaisms, using two measures. The first measured a delta of the last (known) and the first (known) relative frequency of a word, and the second measure checked the correlation between the relative frequency of a word to its average frequency. Both measures produced good, and complementary results, in manual evaluation: the 93--99\% accuracy among the top 100 and bottom 100 ranked lists indicating that for these two change classes, frequency is a good indicator. 
\citet{morsy2016accounting} framed their work in document retrieval and document similarity across time, and made use of link information and frequency information to implicitly account for language change. Static and dynamic topic models were adapted to include links (that is, citations links)  to capture lexical replacement. The dynamic topic models also allowed for a time specific term-topic matrices where, instead of smooth transitions, a term's transition between two time periods was affected by the difference in normalized frequency of the term. 
A modern document was used as a query to find other related documents from earlier time points. The results showed that incorporating link information was useful for document similarity search, and more useful the further apart in time the documents were situated. The frequency information was only sometimes useful; in practice, few terms have a high difference in normalized frequency in adjacent time periods and thus, the transitions of the dynamic topic model resembles smooth transitions.

	 \citet{WordsAreMalleable} offered an alternative to change detection over time, and also studied detection of synchronic variation over viewpoints.
 They relied on Word2Vec (Skip-Gram, 300 dimensions and window size of 10) for creating word embeddings and offered an addition to the linear mapping proposed by \citet{DiachronicWordEmb} for projecting the embeddings onto the same space. Their method takes the neighboring words into account. The authors presented results for pure linear mapping, only neighbor-based approach, and a combination of both.  They studied the two laws proposed by \citet{DiachronicWordEmb} and found that the law of conformity holds across different political viewpoints while the law of innovation does not hold. In addition, they contributed with a law of their own and showed that abstract words are more likely to exhibit different senses according to viewpoint. 
 \citet{fiser-ljubesic-2018} also study synchronic variation and applied a distributional method (word2vec) to investigate the sense distribution of 200 Slovene lemmas in a Slovene social-media dataset (200 MW). They compared to a large corpus (1.2 GW) of published written Standard Slovene, under the assumption that the social-media texts would be ``early adopters'' of incipient semantic changes. They noted a mix of presumably real semantic shifts and register-based or genre-based differences in word sense distributions. Given the authors' explicit aims of providing input for lexicon compilers, this difference is not as important as it would be in a purer historical linguistic research setting, where synchronic polysemy should be kept distinct from diachronic sense change.
	
        \subsection{Sense-Differentiated Change Detection}\label{sec:sdcd}
        
        The methods presented thus far do not currently allow us to recover the senses 
        and therefore, no way of detecting \emph{what} changed. Most methods show the most similar terms to the changing word to illustrate what happens. However, the most similar terms will only represent the dominant sense and not reflect changes among the other senses or capture stable parts of a word. 
        In this section, we review methods that first partition the information concerning one word on the basis of sense information. 
        There are several methods for detecting senses; some rely on word sense induction (also called discrimination)       ; some use topic models
        ;  and some rely on a general clustering mechanism. 
        \footnote{The work of \citet{Tang2016} is presented in Section \ref{sec:WLSC}, under Entropy-based methods as it is a follow up on \citet{Tang2013} where the entropy-based method is presented.} A few of these attempt to track senses over multiple time spans.          We will start by reviewing the topic-modeling and move to word sense induction methods.

\subsubsection{Topic-based models}
	Common for all topic-based models is that the topics are interpreted as senses. With the exception of \citet{Wijaya} that partition topics, no alignment is made between topics to allow following diachronic progression of a sense.
	Topics are not in a 1--1 correspondence to word senses \citep{blei-lafferty-2006,TopicsOverTime} and hence new induction methods aim at inferring sense and topic information jointly \citep{wang2015sense}. 

\paragraph{Detecting novel word senses}    
In their work, \citet{NovelSenses} used topics to represent word senses and performed implicit word sense induction by means of LDA. In particular, a non-parametric topic model called Hierarchical Dirichlet Process \citep{Teh} was shown to provide the best results on the word sense induction task for the Semeval-2010 shared task.  The number of topics was detected rather than pre-defined for each target word, which is beneficial when detecting word senses because all words have different number of senses in different datasets.  
The novel sense detection task was defined with the goal of detecting one or more senses assigned to a target word \textit{w} in a modern corpus that are not assigned to \textit{w} in an older reference corpus. For each target word \textit{w}, all contexts from both corpora are placed in one document $D_w$; the sentence with the target word, one sentence before and one after are used as a context. 

First, topic modeling was applied to the document $D_w$ and all topics were pooled (consisting of topics from both the modern and the reference corpora). 
 Second, each instance of a target word $w$ in the two corpora was assigned a topic. 
 Finally, if a topic was assigned to word instances in the latter corpus but not in the former, then it was considered novel. A novelty score was proposed which considers the difference in probability for topic assignments normalized by a maximum likelihood estimate. The novelty score was high if the sense was more likely in the modern corpus and relatively unlikely in the reference corpus. 

In the work by \citet{NovelSenses}, the written parts of the BNC reference corpus were chosen as the reference corpus, and the second, modern corpus was a random sample of the 2007 ukWaC Web corpus \citep{ukWaC}. Ten words were chosen for a more detailed examination, half of which were manually assessed to have experienced change while the other half had remained stable over the investigated time span. When ranked according to the novelty score, the five words with novel senses (hence forth novel words) were ranked in positions 1, 2, 3, 5, and 8. When repeating the experiment with frequency ratios, the novel words were ranked in positions 1, 2, 6, 9, and 10, indicating that pure frequency is a worse indicator than the novelty score in the case of two corpora that are wide apart in time and content.

In follow-up work, \citet{Cook13alexicographic} proposed  a relevance score that incorporates a set of topically relevant keywords for expected topics of the novel senses, with the main aim to improve the filtering of non-relevant novel senses. In this work, two sub-corpora  of the GIGAWORD corpus for  the  years  1995 and  2008 are used. The experiments in \citet{Cook13alexicographic} differ from that of \citet{NovelSenses}, in that instead of using a pre-defined set of evaluation words. \citet{Cook13alexicographic} used the top 10 words of the novelty score, the rank sum score, and a random selection for further investigation. The evaluation was staged in a lexicography setting and evaluated by a professional lexicographer. Half of the words found using the novelty score had no change in usage or sense. From the words found using the rank sum scores, all words were of interest. From the random chosen words only 3 words were of interest. The interesting cases were then  analyzed by a lexicographer and found to belong to two different classes; having a novel sense (4 plus one of the randomly chosen ones) or in need of a tweak/broadening (9 plus two of the random ones). 

A larger evaluation was performed by \citet{cook:Coling} where two corpus pairs were used, the BNC/ukWaC and the SiBol/Port corpus (that consists of a set of British newspapers, similar in theme and topics, from 1993 and 2010), with 7 and 13 words with novel senses respectively, and a signfincantly larger set of distractors. 
Two additional novelty scores were used, one based on the difference in frequency and one on the log-likelihood ratio of an induced sense in the two corpora. 
There were two relevance scores using topically relevant keywords; the first used an automatically chosen set of 1000 words, and the second used a manually selected subset of the 1000 words. The relevance score was based on the sum of the probability that a relevant word is seen, given an induced sense. 
In addition, the rank sum score was used. 

For the BNC/ukWaC pair, the novelty scores were outperformed by the frequency ratio baseline and the relevance score performed well on its own. %
For the SiBol/Port pair, 
the relevance score with the manually chosen subset outperformed the automatically derived set. 
The rank sum scores performed the best for both corpora pairs. %

Though it was not suggested by the authors in this series of papers \cite{NovelSenses,Cook13alexicographic,cook:Coling}, the method could be used to find the inverse of novelty as well. If a topic is assigned to instances in the reference corpus but not in the second corpus, then the sense can be considered outdated or, at least, dormant. 
Overall, the method proposes the use of topic modeling for word sense induction and a simple method for detecting novel senses in two separate corpora, both by using novelty scores and by incorporating topical knowledge. The senses were, however, not tracked; the exact same sense is expected to be found in both the reference and the modern corpus. 
Assume for example that there is a sense $s_i$ in the reference corpus that does not have a match in the modern corpus, and a sense $s_j$ that has a match in the modern but not in the reference corpus. If $s_i$ is similar to $s_j$, then the two senses could be linked, and possibly considered broadening or narrowing of each other. The difference in $s_i$ and $s_j$ could also be a consequence of random noise. 
By not considering the linking of topics, and only two time points, the complexity was significantly reduced. Drawing on work like that proposed by \citet{Mei_2005}, it remains for future work to track the topics over multiple time periods so additional change types can be detected beyond novel senses.

\paragraph{Clustering and tracking topics}
The work of \citet{Wijaya} addressed some of the weaknesses of the novel sense detection methods, by targeting automatic tracking of word senses over time, where word senses were derived using topic modeling. 

The experiments were conducted on Google Ngram data where 5-grams were chosen in such a way that the target term $w$ was the middle (third) word.
A document $D^{i}_w$ was created for each year $i$ consisting of all 5-grams where $w$ was the third word. Then these documents were clustered using two different methods. 
The first experiment made use of the K-means clustering algorithm and 
the second experiment made use of the Topic-Over-Time algorithm  \citep{TopicsOverTime}, an LDA-like topic model. 
In the \textit{K-means}  experiment, topics were considered to have changed if two consecutive years were assigned to different clusters.  To reduce noise, only clusters that had a consecutive run for more than three years were chosen. Let us assume that cluster 1 contains documents $\{ D^{1}_w, D^{2}_w, D^{3}_w, D^{4}_w  \}$ and cluster 2 contains documents $\{  D^{5}_w, D^{6}_w, D^{7}_w, D^{8}_w  \}$. Then years 4--5 was the change period and the top (tf-idf) words for year 4 and year 5 that represent cluster 1 and cluster 2 were used to represent the different meanings. Clusters that contain years that were not consecutive were removed, meaning that the clustering was used as a partitioning of time periods. In both algorithms, the number of clusters and topics was predetermined and does not relate to the number of senses of the target word. 

For the Topic-Over-Time clustering, two topics were created and the algorithm outputs a temporal distribution for each topic. At each time point, there was only one document. While not directly specified, the strength of a topic $i$, for $i = 1 , 2$ for a time period was likely the assignment of topic $i$ to the document at time $j$. When the most probable topic for a document changes, so does the sense of the word target word $w$.~\footnote{Using the K-means algorithm on documents does not represent  a fully sense-differentiated method. The Topic-Over-Time method represents only two senses active at the same time, and those are constant over time. These two senses correspond to having one representation for two different major senses over different times, where one hands over to the other. Still, the we have chosen to categorize the method among the sense-differentiated methods.  }

A few different words are analyzed; two words changed their dominant sense, \emph{gay} and \emph{awful}. Two words added a sense without losing their previously dominant sense,  \emph{mouse} and \emph{king}, where the latter also became a reference to Martin Luther King. In addition, the authors tested the method for changes to a named entity, Iran's change from monarchy to republic, and John F. Kennedy's and Bill Clinton's transitions from senator to president. Both algorithms captured the time of change, either by a change in cluster or topic distribution.

Adjectives do not seem well suited for the method as their meaning was not well captured by topic models. This might be because topic modeling is not optimal for capturing word senses \citep{TopicModelsWSD}. 
In general, the work presented by \citet{Wijaya} was preliminary but it was the first paper to provide an automatic method for working with more than one sense of a word to find out \emph{what} happened in addition to \emph{when}. There was no proper comparison between the different algorithms to indicate which method performs better or to quantify the results. Two questions remain unanswered. One is, how many of the 20 clusters in K are reasonable? Another is, how often, on average, do we see a change in cluster allocation for the K-means clustering. Nevertheless, the overall methodology of using clustering to associate different topics or documents with each other could be a promising direction.

\paragraph{Dynamic topic models}

\citet{Frerman2016} proposed a dynamic topic model, called SCAN, that differs from the above in several aspects. First, the topic models in their proposal are not independently created for each period, but rely on the adjacent time period. Implicitly, there is a tracking of senses over multiple time periods. Second, each topic can exhibit change over time, to capture subtle changes within a sense. Like the Topics-over-Time algorithm, this dynamic Bayesian model produces a set of fixed topics with a time distribution to show their probability over time. It also allows for  changes over time within each topic as well. An example was given to highlight the importance of allowing senses to change.  The word \emph{mouse} changed, from the 1970s, where words like \emph{cable, ball, mousepad} were important, to \emph{optical, laser, usb} which are important today. All the while both representations stood for the computer device sense of \emph{mouse}.

The DATE corpus, spanning the period 1700--2010, was used for the experiments. 
The corpus was tokenized, lemmatized and part-of-speech tagged, and stopwords and function words were removed. All contexts around a target word \textit{w} from a year \textit{t} were placed in one document, and a time span was 20 years. A context window of size $\pm 5$ was used, resulting in what can be seen as a 11-gram with the target word in the middle, as the 6th word. For two out of three experiments, the number of senses was set to 8. In the third experiment, the number of senses was set to 4. 

The first experiment was a careful analysis of four positive words, namely \textit{band, power, transport} and \textit{bank}. 
For each word and topic number ($1 \ldots 8 $), there were (at most) 16 different topical representations, one per time period. 
On average, 1.2 words were exchanged, a number that was controlled by the precision parameter. No quantification of this number (in relation to the precision parameter, or on its own) was given. The words that stayed among the top-10 did, however, move in rank over time, which signified change without the words being exchanged. 

The second experiment considered novel sense detection \cite{NovelSenses}
and borrowed its evaluation technique from \citet{Mitra2015NLE} and its relevance ranking from \citet{cook:Coling}.  The results for eight time pairs, with a reference and a target time, were presented. 

In this experiment, the number of senses was set to 4. As a baseline, the same model was used to learn topics independently (i.e.,  without the dependency on previous time periods) and was called SCAN-NOT. For this, the topics were matched across time periods using the Jensen-Shannon divergence measure, the topics with the lowest JS divergence were assigned to the same topic number. There was no lower threshold so topics that were very different, but still had the lowest divergence could be assigned to the same topic number. 
Novelty scores were calculated using the relevance score to determine when a topic represents a novel sense. A total of 200 words were identified as sense birth candidates. For the 8 time pairs, SCAN performed better than SCAN-NOT in 6 cases, with a precision score of roughly 0.4--0.65\footnote{These values were read from the bar plot presented in the paper and are not exact.}.

The final experiment\footnote{The authors presented a fourth experiment on the SemEval-2015 DTE task for identifying when a piece of text was written, which we have not presented here.} related to word meaning change and made use of the test set presented by \citet{Gulordava}. The test set consists of  100 words annotated on a 4-point scale, from no change to significant change. The novelty score (as defined by \citealp{cook:Coling}) was calculated on the same 100 words, with 1960's compared to 1990's, and 8 senses per word. The result was the Spearman's rank correlation between the novelty scores and the human ratings from the test set. The correlation score for SCAN was 0.377, as compared to 0.386 reported by \citet{Gulordava} on a different, and larger training set. The SCAN-NOT (0.255) and frequency baseline (0.325) performed worse than SCAN.

The study leaves open questions. For example, the authors did not properly argue for the choice of 8 topics per word, and from the experiments it seems like a large number; for the word \emph{power} three senses were identified; `institutional power', `mental power' and `power as supply of energy'. These were distributed over 4, 3 and 1 topics respectively. What would happen with a lower number of topics? 
The time span of 311 years was partitioned into 8 time periods, which significantly reduced complexity of evaluation. How the method performs with smaller time spans and more time periods remains to be evaluated. 

While novelty of senses was evaluated in detail, there was no discussion of how to differentiate change types or how the method would perform on control words. For the small, in-depth evaluation presented on four words, we saw that all 8 associated topics change\footnote{Change was measured in terms of topical strength, the overlap of the top-10 words between adjacent time periods was not specified.} over time for each word. For example, the `river bank' sense of \emph{bank} should reasonably exhibit a stable behavior, not change so radically over time, to allow the distinction of a stable sense from a changing sense. 
The evaluation of change in individual topics also remains for future work. Is the change in top-10 words or the change in probability of the same set of words over time reasonable for a sense?

The SCAN-method represents an interesting approach that contains most of the necessary components for studying semantic change. Topics were modeled (for individual time periods but with a dependence on previous times) and automatically linked over time, and  were  themselves allowed gradual change. This could enable tracking of individual senses for a word and their rise and fall; it could link them according to concepts and separate the stable senses from the changing ones. 
We highly encourage additional studies into these possibilities.

    \subsubsection{WSI-based}
    	WSI-based models were utilized by \citet{WSE-Mitra,Mitra2015NLE}, \citet{tahmasebi2013models}, and \citet{Tahmasebi-RANLP17} to reveal complex relations between a word's senses by (a) modeling senses \emph{per se} using WSI; and (b) aligning senses over time. The models allow us to identify individual senses at different periods in time and \citet{Tahmasebi-RANLP17} also merges senses into linguistically motivated clusters.

\paragraph{Chinese Whispers}
The work of \citet{WSE-Mitra} was followed up by \citet{Mitra2015NLE}, which presented a more comprehensive analysis. In this review, we will refer to the 2015 work, which almost completely covers the earlier work. 

The aim of the experiments was to track senses over time and to identify if the sense changes were due to birth (novel senses), death (disappearing senses), join (broadening of senses by two senses joining into one), and split (narrowing of a sense by a sense splitting into two). The core part of an approach like this is the method for finding sense clusters. In this work, the method used for detecting senses was the Chinese whispers algorithm \citep{Biemann_2006}. It is based on clustering a co-occurrence graph. For each word, a set of 1000 to features are kept, where features are derived from bigram relations. A pair of words are linked in the graph if they share a sufficient amount of features. The local graph is clustered by starting with all nodes as individual clusters and then merged in a non-deterministic manner, to maximize edge weights of the clusters. To overcome some of the randomness, the procedure is run multiple times and the results are pooled.

Once the clusters are in place, the tracking begins. For each two adjacent time periods, the set of clusters for a word $w$ are compared and the word overlap between any two clusters is measured. To detect birth or join, the overlap is divided by the size of the cluster in the newer period and, inversely, the older period for death and split. A set of criteria determine to which clas the clusters belong.

Two datasets were used in the experiments, Google Books Ngrams (1520--2008) and Twitter (2012--2013). The former dataset was split into eight periods where the first spans 1520--1908 and the last spans 2006--2008. The aim was to have roughly equal amounts of text in each time span. 
The clustering was applied in each time period separately, and compared to all subsequent time periods (and between Google Ngram and Twitter for a cross-media analysis). 
A set of candidate births (ranged from roughly 400 to 4200) were detected between each time span. These changes are considered stable if, for example, a novel sense $s$ that was detected in $t_2$ compared to $t_1$ was also novel in $t_3$ compared to $t_1$. 

The evaluation was performed using two methods, one manually and one automatic. For the manual evaluation, the time period 1909--1953 is compared to 2002--2005. A set of 48 random birth words and 21 random split/join words were inspected manually. The accuracy was 60\% for birth cases and 57\% for split/join.   A set of 50 births were evaluated with respect to Twitter and Google Ngram, out of which 70\% were correct (between datasets no joins or splits were found). 

The automatic evaluation is done with respect to WordNet where clusters for a word $w$ are mapped to a synset of $w$. The method makes use of a synchronic sense repository for detecting sense changes. The mapping is done on the basis of the words in each cluster and their presence as synset members. Roughly half of the clusters are mapped to a synset, but no formal evaluation is conducted. A birth is a success if a cluster $s_{new}$ gets assigned a WordNet synset ID that is not assigned to any of the word's clusters in the earlier period. A split is a success if the two clusters in the new time period  have different synset IDs ($s_{new1} \neq s_{new2} $) and one of them is the same as the old cluster ($s_{new i} = s_{old}$, for $i=1,2$). The join success criteria is analogous to the split criteria, where the new and old time period have swapped places.  For the manual evaluation, the period 1909--1953 was compared to all succeeding periods. While average accuracies were not given, the histogram showed values ranging from roughly 48\% to 62\% for births, from 38\% to 53\% for splits, and from 30\% to 64\% for joins. 

The method does not track senses over multiple time periods; the tracking is done pairwise. This means that the functionality is currently not in place to track a novel sense that is later joined with another. 
While there is a filtering that requires that a novel sense should still be novel in the next time period, the tracking is not done over the entire time period.

\paragraph{Curvature clustering}
The work of \citet{tahmasebi2013models} and \citet{Tahmasebi-RANLP17} has a long-standing basis in (manual) studies related to  diachronic conceptual change on the basis of the curvature clustering algorithm. The aim is to track word sense clusters over time, for individual senses of each word, and to group senses into semantically coherent clusters. Related senses should be grouped together, while unrelated senses should be kept apart. 

The basis of this line of study is the word sense clusters, that rely on the curvature clustering algorithm \cite{Dorow04usingcurvature} applicable to nouns and noun phrases in coordination. 
\citet{Dorow04usingcurvature} investigated the quality of the clusters on WordNet for modern data (British National Corpus) and \citet{Tahmasebi2013-IJDL} evaluated the quality with respect to historical data. The quality of the clusters remained high despite the higher amount of OCR errors, but the amount of extracted clusters dropped with higher error rates. 
The experiments were conducted on the (London) Times Archive and the New York Times annotated corpus, on yearly sub-corpora. The resulting dataset spanned 1785--1985 and 1987--2007. %

Then the cluster sets for a target word \textit{w} were compared over subsequent years. The comparison was done using a modified Jaccard similarity (to boost similarity between clusters of largely different sizes but with high overlaps) and a WordNet-based similarity measure based on the \citet{Lin_1998} measure. 
In the first phase, clusters that were similar enough to be considered the same over time (including some random noise) were grouped. These groupings correspond to stable senses over an arbitrary time span. In the next phase, these groupings were compared across all time periods. This two-step procedure was used to reduce the complexity, as otherwise, the possible transitions between clusters grow exponentially with the number of clusters and time periods. 
After these two first steps, there were a set of linked senses over time for a target word. As a final step, the individually linked senses were grouped into semantically coherent groups 
, while unrelated senses belonged to different groups. 

The method allows for the detection of broadening and narrowing, splitting and merging senses, novel related and novel unrelated (\eg neologisms) senses, and stable senses. Each change event was monitored individually, hence a word could first have a novel sense that later changed by means of, for example, broadening. These were then considered two separate change events. The stable senses could belong to two different categories, those words that had no change events and were stable over the entire time span and those that experienced change in another sense. 

The test set consisted of 35 change events corresponding to 23 words, and 26 non-change events. Eleven of these corresponded to stable words without other change events and the remainder corresponded to words that had change events related to their other senses. In addition,
the authors also evaluated the time delay with which the change was found with respect to both a ground truth dataset and to the first available cluster representing a given sense or change. 
On average, 95\% of all senses and changes were found among the clusters, showing the upper limit for the word sense induction method on the dataset. 
Eighty-four percent of the change events could be discriminated and correctly identified. 
Only related, novel senses could not be found properly, most likely due to little or no word overlap in the contexts. 

The average time delay was presented as a time span between two time points. 
The first represents the manually chosen outside world (and can be the time of invention or the first attested use of a word sense) but needs not to be valid for this specific dataset. The second represents the time the (automatic) word sense induction method can detect evidence of a sense or change. If the gap between these two time points is large, there is either little evidence in the datasets, or the WSI method was unable to detect the sense. 
The true time delay lies between these two points. 
For detected senses and changes, the time delay is on average 6.3--28.7 years. For the change events that can be discriminated and correctly identified, the time delay is slightly higher, 9.9 -- 32.2. In particular, existing senses of words with change events have a time delay of 11.7 -- 59.0, while the corresponding number for words without change events is much lower, 2.7--20.5 years. These delays can be compared to those present presented by \citet{basilediachronic} who found, on average, a time delay of 38 years for change in the dominant sense. 
This speaks to the fact that words are unlikely to change their meaning if they are frequently in use.

The strength of the method is the possibility of tracking senses on an individual basis; and to allow for certain parts of a word to stay stable while other parts change independently of each other. The food sense of an \textit{apple} does not disappear because the company Apple is the more popularly used sense.  All senses are tracked over each year, which increases the complexity but keeps a fairly high granularity for change detection. The authors did not filter any results and hence presented no precision.

\subsubsection{Aligned corpora}

The work conducted by \citet{Bamman} sought to track the rise and fall of Latin word senses over 2000 years. They used 
two aligned corpora in different languages for translation of words to help approximate the senses of the word. 
The number of different translations in language B will provide a probable guess on how many different senses are valid for the word in language A. The translation mechanism also helps to determine the frequency with which the instances of the target word are assigned to the senses; the more often the target word is translated to word $i$ in language B, the more often the sense $i$ is assigned to the target word in language A. 

The results clearly showed that sense variations could be measured over time and pointed to a change in the predominant sense over time for five chosen terms. 
The method is far more beneficial for studying words and their meanings over time than studies based on word frequency. 
However, it is limited as it requires a translated corpus to train the word sense disambiguation classifier. In addition, it does not allow the senses to be aligned over time to follow the evolution of senses and their relations. 

\subsection{Comparison}

Finally, Table~\ref{tab:datasets} gives an overview of the datasets used, and Table~\ref{tab:compOfWSCmethods} provides a summary with respect to the most important aspects and differences of the studies reviewed in this section.

   \begin{table}[bht]
   \small
\centering
\caption{Datasets used for diachronic conceptual change detection.  Non-English *}\label{tab:datasets}
\begin{tabular}{ll}
\toprule
\citet{sagi2009semantic} & Helsinki corpus\\
\citet{Gulordava} & Google Ngram\\
\citet{Wijaya} & Google Ngram\\
\citet{NovelSenses} & British National Corpus (BNC), ukWaC\\
\citet{Cook13alexicographic} & Gigawords corpus\\
\citet{cook:Coling} & BNC, ukWaC, Sibol/Port\\
 \citet{Mihalcea_wordepochdis} & Google books\\
 \citet{basilediachronic} & Google Ngram (Italian)\\
\citet{Tang2013,Tang2016}* & Chinese People's Daily\\
\citet{TempAnal-NeuralLangModel} & Google Ngram\\
\citet{kulkarni2015statistically} & Google Ngram, Twitter, Amazon movie reviews\\
\citet{Mitra2015NLE}& Google Ngram, Twitter\\
\citet{DiachronicWordEmb} & COHA, Google Ngram\\ 
\citet{Eger-SemChange}* & COHA, S\"uddeutsche Zeitung, PL\tablefootnote{Patrologiae cursus completus: Series latina}\\
\citet{WordsAreMalleable} & New York Times Annotated Corpus, Hansard\\ 
\citet{rodda2016panta}* & Thesaurus Linguae Graecae\\
\citet{Frerman2016} & DATE corpus\\ 
\citet{JapnLoanwords} & Wikipedia (English and Japanese)\\
\citet{KahmannNH17} & Guardian (non-public) \\
\citet{Tahmasebi-RANLP17} & Times Archive, New York Times Annotated Corpus\\
\citet{bamler17} & Google Books Ngrams, State of the Union addresses, Twitter \\
\citet{Yao-2018} & New York Times (non-public)\\
\citet{RudolphB18-dynamicEmbforLangEvo} & ACM abstracts, ML papers ArXiv, U.S. Senate speech\\
\bottomrule
\end{tabular}
\end{table}     

\changetext{0pt}{13em}{}{}{}%
\begin{landscape}
\begin{table}

\scriptsize
\centering
 \begin{minipage}{\textwidth}
 \begin{adjustwidth}{-1in}{-1in}%
\caption[Comparison for methods in WSC]{Comparison of methods for diachronic conceptual change detection} \label{tab:compOfWSCmethods}
\begin{tabular}{lc@{}cc@{}cc@{}cc@{}c@{}ccc}
&\multicolumn{2}{c}{prechosen} & top & entity &eval. method & \multicolumn{2}{c}{time} & \# classes & classes & \multicolumn{2}{c}{modes}\\
& \# pos & \# neg & & \scriptsize{(S)ingle/}   & \scriptsize{(M)anual/} & span & \# points & & & \multicolumn{2}{c}{\scriptsize{time / sense}}\\
&  &  & & \scriptsize{(P)airs}   & \scriptsize{(A)utomatic} &  &  & & & \multicolumn{2}{c}{\scriptsize{aware / diff}}\\
\midrule
\scriptsize{\citet{sagi2009semantic}} & 4 & 0 & &S& M&  569y & 4 & 2 & broad./narrow.& no&no  \\
\scriptsize{\citet{Gulordava}} & 0&0 & 100\tablefootnote{\scriptsize{ 100 randomly chosen words.}} & S & M & 40y & 2 & 1 & change & no &no\\
\scriptsize{\citet{Tang2013}}& 33 & 12& & S & M & 59 & 59 & 3 & B/N/novel/change\tablefootnote{\scriptsize{ Broadening is divided into two subclasses with metaphoric and metonymic change.}} & no & no\\
\scriptsize{\citet{TempAnal-NeuralLangModel}} & 0 & 0& 10/10\tablefootnote{\scriptsize{ 10 most positive and 10 most negative words are evaluated.}} & S/P\tablefootnote{\scriptsize{Two words \textit{cell} and \textit{gay} are evaluated in pairs, where the paired words are related to their different senses, \eg \textit{closet} and \textit{phone} for the word \textit{cell}.}} & M & 110 & 110 & 1 & change & yes\tablefootnote{\scriptsize{ They do not explicitly evaluate time.}} & no\\
\scriptsize{\citet{kulkarni2015statistically}}& 20 & 0 &20\tablefootnote{\scriptsize{ Top 20 words per method were evaluated.}} & S & M/A & 105y/12y/2y & 21/13/24 & 1 & change & yes&no\\
\scriptsize{\citet{DiachronicWordEmb}} & 28 & 0 & 10\tablefootnote{\scriptsize{ Top 10 words per method were evaluated.}} & S/P & M  & 200/190 & 20& 1 & change & no&no\\ 
\scriptsize{\citet{rodda2016panta}} & 0 & 0 & 50  & S & M & 1200y & 2 & 1 & change& no & no\\
\scriptsize{\citet{Eger-SemChange}} & 0 & 0 & 21\tablefootnote{\scriptsize{Top 9 plus two more from position 16 and 36 and bottom 10 words are chosen.} } & S/P & M & 200/190 & 20/19 & 1 & change & no & no\\
\scriptsize{\citet{basilediachronic}} & 40 & 0 & & S & M & 170& 17 & 1 & change & yes & no\\
\scriptsize{\citet{WordsAreMalleable}} & 24 & 0& 5/5\tablefootnote{\scriptsize{ Top 5 words per method and bottom 5 (most stable) words were evaluated.}} &S & M & 20/11& 2/2& 1 & change & no & no\\
\scriptsize{\citet{JapnLoanwords} }& 10 & 0 & 100/20\tablefootnote{\scriptsize{ Top 100 positive and 20 negative words were evaluated.}} & S/P & M & -\tablefootnote{\scriptsize{Method aimed to detect difference in loanwords, and hence time span was not considered, instead two snapshots of Wikipedia were used.}} & 2 & 1 & change & no & no\\
\scriptsize{\citet{KahmannNH17}} & 4 & 0 & & S & M & $\leq$ 1\tablefootnote{\scriptsize{11 months of data is used.} } & 48 & 1\tablefootnote{\scriptsize{ While three classes of change is evaluated in the experiment on artificial data, only volatility was considered in the experiments on real data.}} & change& no & no\\
\scriptsize{\citet{bamler17}} & 6 & 0 & 10 & S/P& M\tablefootnote{\scriptsize{ There is no automatic evaluation of change, but of smoothness of the dynamic embeddings.}} & 209/230/7 & 209/230/21 & 1 & change & no & no\\
\scriptsize{\citet{Yao-2018}} & 4/1888\tablefootnote{\scriptsize{ 4 words were evaluated manually, 1888 were evaluated automatically, proportion of pos./neg. is unclear.}} & 0 & & S & M/A & 27 & 27 & 1 & change & no & no\\
\midrule
\scriptsize{\citet{Wijaya}} & 4 & 2 & & S & M & ~500\tablefootnote{\scriptsize{The authors use the full Google Books Ngrams corpus without reference to the exact dates of inclusion.} } & ~500 & 2\tablefootnote{\scriptsize{ Two change classes are used for analysis but the algorithm does not differentiate between the kinds.}}& change\ novel & yes & yes\tablefootnote{\scriptsize{ The Topic-over-Time algorithm provides two topics, interpreted as senses, that are valid at the same time. Mostly these are one sense that hands over to another, but since both topics can be active at the same time, we consider this to fall into sense differentiated methods.} }\\
\scriptsize{\citet{NovelSenses}} & 5 & 5 & & S & M & 43 y & 2 & 1 &  novel &no&yes\\
\scriptsize{\citet{Cook13alexicographic}} & 0 & 0& 30 & S & M & 14 & 2 & 1 & novel & no & yes\\
\scriptsize{\citet{cook:Coling}} & 7/13 & 50/164 & & S & M & 43y/17y & 2/2 & 1& novel & no&yes\\
\scriptsize{\citet{Mitra2015NLE}}\tablefootnote{\scriptsize{ The analysis is extended with a Twitter corpus from \citet{WSE-Mitra}.}} & 0 & 0 & 69/50& S & M/A & 488/2 & 8/2& 3& split/join/novel\tablefootnote{\scriptsize{ The class 'death' is briefly discussed as well.}} & no&yes\\
\scriptsize{\citet{Frerman2016}} & 4 & 0 & 200 & S & M/A & 311 & 16 & 2 & change/novel & no & yes\\ 
\scriptsize{\citet{Tang2016}}\tablefootnote{\scriptsize{ \citet{Tang2013} performed a smaller set of experiments on the same data using the same change classes, but without the sense differentiation.}} & 197 & 0 &  & S & M&  59 & 59 &  6& B/N/novel/change\tablefootnote{\scriptsize{ Broadening is divided into two subclasses with metaphoric and metonymic change.}} & no & yes\\
\scriptsize{\citet{Tahmasebi-RANLP17}} & 35& 25 & & S & M& 222y & 221 & 4& novel,B/N,stable & yes&yes\\ 
\bottomrule
\end{tabular}
\end{adjustwidth}%
\end{minipage}
\end{table}
\end{landscape}

\changetext{0pt}{-13em}{}{}{}%

\section{Computational Modeling of Diachronic Word Replacement}\label{sec:dwr}

In general, one can distinguish the following types of diachronic replacements: 

(1) Lexical replacements relate to words from any part of speech and its detection requires sense information. Words may have different sets of senses at different times and some of the senses can be replaced by others. Examples include 
``foolish'' that replaced ``nice'' for the foolish sense of the latter, and ``cool'' that replaced ``relaxed.''\footnote{The latter replacement is seen as a synchronic variation as both words ``cool'' and ``relaxed'' are used in different populations to mean the same thing. In the former case, ``nice'' has completely lost its \textit{foolish} sense. } 

(2) Terms that describe the same entity/object at different times and represent different names of that entity/object. For example, \emph{Myanmar} is the current name of \emph{Burma} and both refer to the same object (same identity). Note that an object here needs to be a named entity (i.e., it has identity). Furthermore, multiple names can be used to refer to the same object at the same time, and some names can substitute for others over time. The latter represents the phenomenon of diachronic \textit{named entity change}.

(3) Terms that are instances of the same type that were valid at different times, for example, the names of US presidents. Note that the instances should usually be exclusive at any given time point (i.e., there is only one US president at a given time point). Here, the analogy consists in the fact that the instances are of the same type/concept and not influenced by the attributional similarity of the instances (e.g., whether president George W. Bush was really similar in its character or other attributes to president Bill Clinton). 

(4) The last type is temporal analogs, which are very similar due to shared role, attributes, functions despite time gap, yet do not belong to the other three types. Analogy in general is a cognitive process of transferring information or meaning from a particular subject called the analogue or source to another subject called the target. Temporal analogy could be considered a subtype of analogy because it is a comparison of two subjects that existed in different times based on their similarity or equivalence. One reason for finding analogous terms in different times is providing support for querying document archives.

\begin{figure}[bt]
	\centering

	\includegraphics[width=0.7\textwidth]{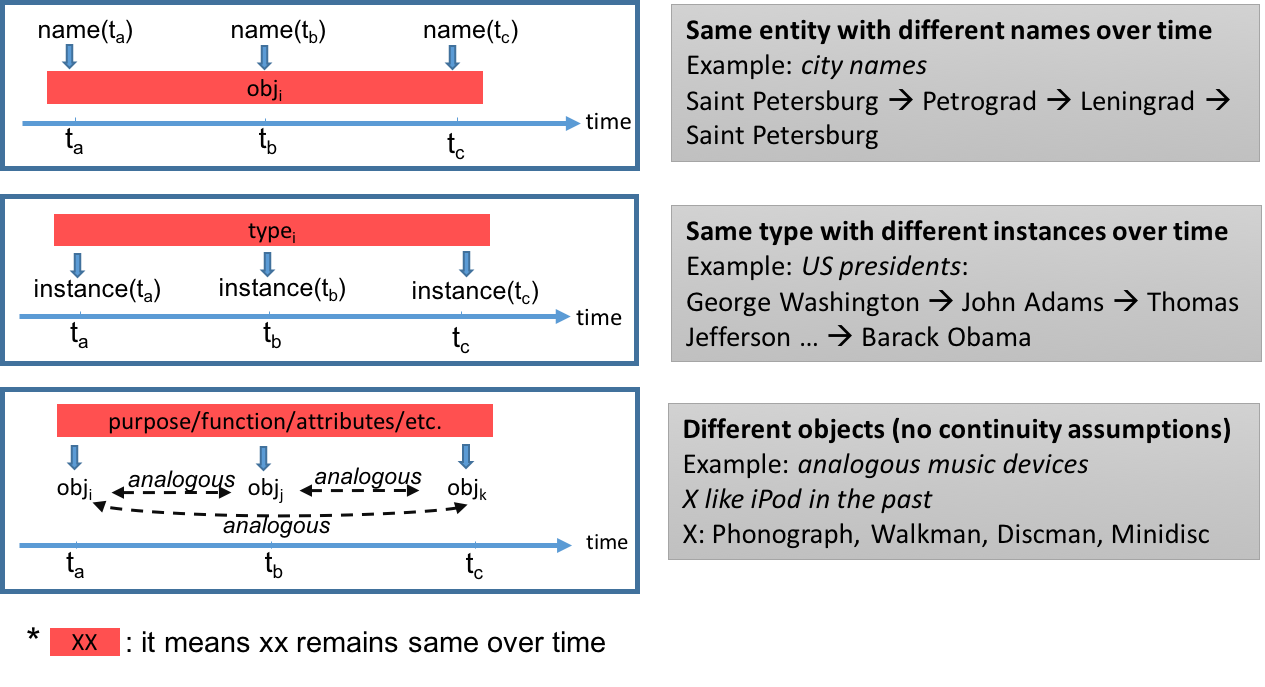}

        \caption[Types of temporal analogs]{Conceptual view of three types ((2), (3) and (4) from the list above) of diachronic replacements.}
	\label{fig:across-time-sim}
\end{figure}

The three latter types (without lexical replacement) are conceptually depicted in Fig. \ref{fig:across-time-sim}. Note that the ability to find diachronic replacements has many applications ranging from educational ones, uses as components in larger systems such as search engines or, in general, in NLP pipelines.

Several works on finding diachronic replacements over time are described below. However most of them do not use the sense information of a word, hence effectively treating a word as having one sense (i.e., often its dominant sense). 

We mainly focus on works related to finding replacement types (3) and (4), i.e., named entity replacements and temporal analogs.

\citet{BerberichBSW09} were the first to propose reformulating a query into terms used in the past hence their motivation was suportng user search experience within document archives. The task was defined as follows: given a query $q = {q_1, q_2,..,q_m}$ formulated using terminology valid at a reference time $R$, identify a query reformulation $q' = {q'_1, q'_2,..,q'_m}$ that paraphrases the same information need using terminology valid at a target time $T$. They measured the degree of relatedness between two terms used at different times through context comparison using co-occurrence statistics. A Hidden Markov model was used for query reformulation; it considered three criteria of a good reformulation: similarity, coherence, and popularity. In particular, the similarity criterion requires that
$q_i$ and $q'_i$ have high degree of across-time semantic similarity, while coherence means that $q'_i$ and $q'_{i-1}$  should co-occur frequently at time $T$ to avoid combining unrelated terms. Finally, $q'_i$ should occur frequently at time $T$ to avoid unlikely query reformulations.
This approach may require a recurrent computation each time a query is submitted because it needs a target time point for the query reformulation.

\citet{SITAC} proposed that semantically identical words (or named entities) used at different time periods could be discovered using association rule mining to associate distinct entities with events. Sentences containing a subject, verb, and object 
are targeted and the verb is interpreted as an event. Two entities are then considered semantically related if their associated event is the same and the event occurs multiple times in a document archive. The temporally related term of a named entity is used for query translation (or reformulation) and results are retrieved appropriately with respect to specified time criteria.

\citet{KanhabuaN10} extracted time-based synonyms of named entities from link anchor texts in Wikipedia articles, using the full article history. Because of the limited time span of Wikipedia, they extended the discovered time of synonyms by using burst detection method on the New York Times Annotated Corpus. Unfortunately, link information, such as anchor text, is rarely available and thus limits the method to hypertext collections. The authors evaluated the precision and recall of the time-based synonyms by measuring precision and recall in the search results rather than directly evaluating the quality of the synonyms found.

\citet{tahmasebi2012neer} proposed a method called \emph{NEER} for discovering different names for the same named entities (\eg \emph{Joseph Ratzinger} and \emph{Pope Benedict XVI}, \emph{Hillary Rodham} and \emph{Hillary Clinton}).
It relied first on detecting the periods that had a high likelihood of name changes and analyzed the contexts during the periods of change to find different temporal co-references of named entities. The key hypothesis was that this approach could capture both the old and the new co-reference in the same context. The underlying assumption was that named entity changes typically occur during a short time span due to special events (e.g., being elected pope, getting married or merging/splitting a company). Co-references were classified as direct and indirect. Direct co-references have some lexical overlap (\eg \emph{President Obama} and \emph{Barack Obama}), while indirect ones lack any lexical overlap (\eg \emph{President} and \emph{Barack Obama}). The proposed method first identified potential change periods via burst detection. Bursts related to an entity were found by retrieving all the documents in the corpus containing the query term, grouping them into monthly bins, and running the burst detection on the relative frequency of the documents in each bin. After NLP processing, the method creates a co-occurrence graph of nouns, noun phrases and named entities from documents mentioning the input entity. The next step collapsed the co-references based on their lexical similarity and merged their contexts into co-reference classes. All terms in the context of a given co-reference class were then considered as candidate indirect co-references. 

\citet{tahmasebi2012neer} conducted experiments on the New York Times dataset (see Section~\ref{Evaluation_diach_replacement}) using 16 distinct entities corresponding to 33 names and 86 co-references (44 indirect and 42 direct). Using a random forest classifier they achieved a precision of
90\% on known time periods and 93\% on found periods. 
The proposed method was later applied for query suggestion in search engines using temporal variants of a query \cite{holzmann2012fokas} and for detecting named entity evolution in the blogosphere \cite{holzmann2015named}.

As typically, there is low overlap between contexts of temporal analogs, solutions that rely on measuring context overlap do not work well. Distributed word representations \citep[\eg][]{Mikolov2013-Distributed} 
can be useful for avoiding problem of low context overlap. Given the representations trained on the distant time periods (typically, one derived from the present documents and another from documents published in the past), matching words across time could be done through transformation. This essentially means aligning relative positions of terms in the vector spaces of different time periods. \citet{zhang2016past} and later \citet{szymanski:2017} 
used a linear transformation matrix for finding translations between word embeddings trained on non-consecutive time periods for detecting temporal analogs. The inherent problem in this kind of approach is the difficulty of finding a large enough training set, given the variety of domains, document genres, and arbitrary time periods for finding temporal analogs. 
A simple solution proposed by \citet{zhang2016past} assumes that frequent and common terms in both the time periods can be easily acquired and used for optimizing the linear transformation matrix. This idea is based on the observation that most frequent words are known to change their semantics across time only to a small degree \cite{DiachronicWordEmb,pagel2007frequency,lieberman2007quantifying}. Initializing word embeddings using embeddings trained on previous time periods \cite{TempAnal-NeuralLangModel} is difficult given the potentially long gaps between the two periods on which the vector spaces were trained. The potential lack of access to data from the intermediate periods can be another problem.
The authors also successfully experimented with using terms that were computationally verified to have undergone little semantic variation across time as training instances for transformation matrix. They did this by comparing sequentially trained word representations from consecutive time periods. Another improvement was the introduction of local approach that relied on transforming automatically selected reference terms for a given query, which are supposed to ground the meaning of the query. Such transformed reference terms were then compared with the reference terms of candidate analogs, which had been generated by the previously described global transformation approach, with a linear transformation matrix. In other words, the global transformation approach was effectively extended with a method that locally constrains a query by transforming selected context terms called reference terms and then compares these terms with ones of candidate analogs. The reference-to-reference term similarity measure relies not only on comparison of transformed vectors but also on comparison of transformed vector differences. The idea behind comparing vector differences was to capture the relation of a query (or a candidate analog) and its reference term.
Three methods were suggested for proposing the reference terms from candidate context terms: PMI, clustering, and hyperonym detection using shallow processing \cite{ohshima2010high} in an attempt to reflect the relevance, diversity, and generality of reference the terms, respectively.
Experiments were done on manually constructed ground truth data consisting of pairs of temporal analogs (see Sec. \ref{Evaluation_diach_replacement} for more details) using precision at different cutoff points and Mean Reciprocal Rank (MRR). The results showed that local approach using reference terms selected from hyperonyms of a query (and of candidate terms) performed the best. The authors also demonstrated that correcting OCR errors by using a simple approach based on word embedding similarity and word frequency greatly enhances the quality of results.

More recently, \citet{Zhang:2017:TAR:3132847.3132917} proposed using a set of transformation matrices based on different hierarchical clusters over the vocabularies in the two time periods. The thinking was that a single linear transformation matrix is insufficient for obtaining good mapping between vector spaces of different periods. However, they found that using a series of matrices that each corresponded to a given hierarchical cluster of terms and aggregating their results performed better.

\citet{Orlikowski2920762} compared number of models that rely on operations on word embeddings using nine different concepts on a corpus of Dutch newspapers from the 1950s and 1980s. Their objective was to test various assumptions. 
Following \citet{kenter2015ad}, the authors assumed the notion of diachronic concept change involving core concept terms and characterizing concept term. Based on that model, the characterizing terms are expected to change over time, while the surface forms of the core terms are assumed to stay the same. The problem of concept change at a given time point is then reduced to the problem of predicting valid characterizing terms for a core concept term, given a respective characterizing term at an earlier point.

All the approaches proposed so far have relied on sense-agnostic solutions, essentially, mixing all the senses (or relying on the dominant sense). A future improvement would be to move into direction of finding analogous terms with respect to their senses or topics/aspects (sometimes called viewpoints). For the example of the latter, consider \emph{Walkman} which corresponds to \emph{iPod} due to their similar function as a ``music device,'' while \emph{PC} can be a reasonable analog when regarding \emph{iPod} as a ``game player''. A queried term, for example, an entity, may contain multiple aspects and the temporal analogs could be different depending on the particular topic/aspect. In this regard, \citet{zhangWSDM} has demonstrated a simple solution for an aspect-based temporal analog retrieval that takes additional term as an input to restrict the meaning of user query to a particular viewpoint or aspect. The proposed solution also utilizes a neural network to realize non-linear term-to-tem mapping. 

Furthermore, all the approaches, with the exception of the work by \citet{tahmasebi2012neer} and \citet{Kaluarachchi}, need clearly specified time periods for comparison. While typically one of the periods represents the present (i.e., the time when a present-day user needs some information), the other can be any period in the past. It is, however, not always feasible to require users to specify specific periods for which temporal analogs need to be output. In many scenarios, it may be assumed that the user wants to know all the analogs from the past, hence, methods that can provide ranked results based on the agglomeration of results collected from different time periods should be also proposed. 

Outputting evidence for automatic explanation of term similarity is a related problem to estimating similarity across time. The approach proposed by \citet{zhang2016towards} relies on providing evidence of terms' similarity over time by outputting explanatory context terms and then extracting sentences that reveal the shared aspects between temporal analogs. For example, for the input query pair \texttt{ipod} and \texttt{walkman}, the pairs of explanatory terms could be \texttt{music}--\texttt{music}, \texttt{device}--\texttt{device}, \texttt{apple}--\texttt{sony}, \texttt{mp3}--\texttt{cassette}, and so on. Note that the input is now the pair of query terms instead of a single term, as it is in the temporal analog retrieval task, and the output is the ranked list of term pairs.
Term pairs are ranked based on their relevance to the input query pair as well as the intra-similarity between the pair elements and their relations to query terms (both similarities are computed after applying transformation).

\section{Linguistic Approaches to Vocabulary Change}\label{sec:lingappr}
The study of how meaning -- in particular, lexical meaning -- is
expressed and manipulated in language is pursued in a number of
scientific disciplines, including psychology, (cultural) anthropology,
history, literature, philosophy, cognitive science, and  in
linguistics and computational linguistics. 

Here, our primary frame of reference is provided
by relevant work in (general) linguistics, being the field offering
the theoretically and empirically best-grounded view on the phenomena
under discussion here. In particular, in studying meaning in language,
linguistics takes a broad cross-linguistic perspective, which is typically
lacking in the other disciplines.

In this article, the linguistic entities in focus are the meanings of \emph{lexical
  items}, corresponding roughly to what is
often called \emph{lexeme} in lexicography
\citep[e.g., ][]{matthews-1974}, i.e., basically an
entry -- a word or multi-word expression -- in a conventional
dictionary. Among the lexical items
we also include proper nouns and function words. Interchangeably with
lexical item we also say ``word'', intending this term also
to apply to multiword expressions. The meaning of a word is often referred to as a \emph{concept}.  Both terms --
``meaning'' and ``concept'' -- unfortunately have many, mutually
incompatible, uses in the literature, and we use the two terms -- sparingly -- 
interchangeably here, with the understanding that neither term is well-defined.

We refer to the combination of a lexical item and a particular
recognized meaning of that lexical item as a \emph{word sense}. Thus,
both \emph{bank} (n) `(a kind of) financial institution' and
\emph{bank} (n) `extended shallow portion of sea floor' are word
senses according to this definition, as are \emph{deer} (n) `a kind of
(game) animal' and \emph{deer} (n) `meat of this animal used as food'.

The relationship between forms and meanings is many-to-many, so
one form may be used to express more than one meaning (\emph{polysemy}), and, conversely, the same meaning can be expressed by more than one form (\emph{synonymy}).

While the form units -- the words -- are comparatively easy to identify
in language, word senses are notoriously difficult to isolate. Much of
the work surveyed here takes a published lexicon as providing the canonical sense set, the
gold standard by which to judge system accuracy. While this is a practical solution for many
purposes, it ignores a host of difficult theoretical and
methodological questions. For the purposes
of this survey, we do not take a stand on precisely how word senses
are defined and identified, but we do note that some of the
approaches represented in the work surveyed  have the potential to
throw light on these questions; see below. 
\subsection{Types of Lexical Change}

To a linguist, the topic of this article would fall under the
rubric of \emph{historical-comparative linguistics} or
\emph{diachronic linguistics}. This is a branch of general linguistics that
concerns itself with how languages change over time and with
uncovering evidence for genetic relations among languages
\citep{Campbell,hl-handbook}.

The phenomena addressed in the works surveyed in this article
(i.e., historical developments in the vocabulary of a language or languages) are
studied by historical linguists under the headings of \emph{lexical (semantic) change}, \emph{semantic change}, \emph{grammaticalization}, and \emph{lexical replacement}.

In linguistic literature, the term lexical change is used in two senses. In the sense used here, it is a general cover term for all kinds of diachronic changes in the vocabulary of a language or languages. The other common usage is a hyponym of this, referring to new words (new forms) entering or leaving the language, i.e., loanwords and
neologisms of various kinds, and obsolescing words, respectively. 

Lexical replacement is used about a lexeme being ousted by another synonymous lexeme over time, as when \emph{adrenaline} is replaced by \emph{epinephrine}. 
A particular form of
lexical replacement which has received a fair amount of attention in
computational linguistics but which is generally not studied at all by
historical linguists is named entity change (see Section~\ref{sec:dwr}).

Semantic change or semantic shift is the normal term for the special case of lexical
change where an existing form aquires or loses a particular meaning,
i.e., increasing or decreasing polysemy
\citep{traugott-dasher,fortson-2003,newman-2016,traugott-2017}. An example are the oft-cited changes whereby an earlier English word for a
particular kind of dog became the general word for `dog', and,
conversely, the earlier general word for `dog' -- whose modern reflex
is \emph{hound} (n) -- is now used for a special kind of dog. In this and other cases where a word seems to have changed from one meaning to another over the course of time, linguists generally assume an intermediate polysemous stage.

The distinction reflects two complementary approaches adopted by linguists to
the study of the lexicon. Lexical items can be studied from the
\emph{onomasiological} point of view, investigating how particular
meanings (or concepts) are expressed in a language. The Princeton
WordNet \citep{fellbaum-1998} is an onomasiologically organized
lexical resource, as is, \eg \emph{ Roget's Thesaurus}
\citep{roget-1852}.  The more common \emph{semasiological} approach
takes linguistic forms -- words and multi-word expressions -- as its
point of departure and investigates which meanings they express. Conventional dictionaries are semasiologically organized.

Studies of semantic change adopt
the semasiological perspective, whereas works on other forms of lexical change generally have an onomasiological focus.

Grammaticalization
\citep{hopper-traugott-1993,heine-kuteva,smith-2011} denotes a
particular kind of semantic change, where content words turn into
function words and ultimately into bound grammatical morphemes. One example is the
French preposition \emph{chez} `at, with', developed from the Latin
noun \emph{casa} `(small) house, cottage'.\footnote{See
  \url{http://www.cnrtl.fr/definition/chez}.}

\subsection{Linguistic Generalizations About Lexical Change} 

General linguistics studies language as
a universal phenomenon, and in this connection an
important concern is the generalization of sets of observed individual
lexical changes into types and classes of changes, valid for human
languages in general. Adding a word sense to the vocabulary of a language  can be accomplished in several different ways: by borrowing, by coining a new word \emph{ex nihilo} (rare) or
by using the word-formation machinery of the language, or finally --
and commonly --  by adding a word sense to an existing lexeme. The latter
can again be achieved by, for example, 
\emph{generalization} or \emph{broadening} (English \emph{dog} `a kind
of dog' $>$ `dog')
and \emph{specialization} or \emph{narrowing}
(English \emph{hound} `dog' $>$ `a kind of dog'). Other types of semantic change have their origin in
\emph{metaphor}, as in the \emph{foot} of a mountain or the
\emph{head} of a state, \emph{metonymy}, for example, the development
where \emph{bead}, a word originally meaning `prayer', acquired its
current meaning from the use of a rosary while praying, and \emph{ellipsis}, as \emph{mobile} and \emph{cell} from \emph{mobile phone} and \emph{cell phone}, respectively. 
 
Finally, a lexeme in one language may add a sense by mirroring a polysemy in another language, a form
of loan translation. For example, the Swedish verb \emph{suga} `to suck' has acquired a recent new sense `to be unpleasant, inferior, etc.' borrowed from English. From this it follows that semantic change typically involves polysemy. Crucially, even cases of seemingly complete sense change in a lexeme are thought to involve an intermediate (unattested) polysemous stage: A $>$ A+B $>$ B, or A $>$ A+b $>$ a+B $>$ B, where A/a and B/b are senses related by some regular mechanism of sense change and caps indicate a dominant sense.

The activities of broadly characterizing and classifying vocabulary
changes overlap significantly with another linguistic subdiscipline,
namely \emph{lexical typology} or \emph{semantic typology}
\citep{k-tamm-2008,riemer-2010,k-tamm-2012,k-tamm-etal-2016}, whose aims are to
elucidate questions such as ``how languages categorize particular
domains (human bodies, kinship relations, colour, motion, perception,
etc.) by means of lexical items, what parameters underlie
categorization, whether languages are completely free to `carve up'
the domains at an infinite and arbitrary number of places or whether
there are limits on this, and whether any categories are universal
(e.g. `relative', `body', or `red')''
\citep[434]{k-tamm-etal-2016}. These questions are 
relevant to classificatory activities, since universal restrictions on or tendencies of 
lexicalization will determine which semantic changes are possible or
likely, as opposed to impossible or unlikely.

A central goal of linguistics is to explain
linguistic phenomena. Hence, a third kind of activity is the
search for enabling factors and, ultimately explanations for the
observed changes and regularities of change formulated on the basis of
broad cross-linguistic comparison. 
In their search for explanations of lexical change, linguists have
proposed some factors that seem to play a role in lexical change,
as (proximal or distal) causes or as enabling or constraining
mechanisms. Material and immaterial culture are almost always
mentioned in this connection. In order to be able to talk about new
objects, phenomena and practices, we need new vocabulary.

Other external factors proposed in the literature are human
physiology and cognition (e.g. in relation to color
vocabulary), the size of the language community, language contact, and
the presence of large numbers of second-language speakers, among others.

\subsection{Historical-comparative linguistics meets computational linguistics?}

While some of the work described in this survey has
not been directly motivated by linguistic research questions, in many cases the authors of these works
indicate the potential usefulness of their results to
linguistics. We believe that computational approaches to lexical and semantic change have the potential to provide a
genuinely novel direction for historical linguistics.  However, this
is not likely to happen without these authors paying more attention to the 
theoretical and methodological assumptions of current historical
linguistics, an awareness sometimes lacking in the work surveyed. For linguists to take notice of this work, it needs to show
awareness of the state of the art of historical linguistics and argue in terms understandable to a linguistic
audience. On the other hand, in our view these computational methods
represent a genuinely novel approach to addressing central questions of
historical linguistics, which linguists must be prepared to assimilate
at least to some extent in order to grasp the implications of the
results. Thus, if these methods are to make an impact on research in historical
linguistics, this will most likely require a conceptual shift for both
parties.

\section{Methodological Issues and Evaluation}\label{sec:method}
\subsection{Evaluation and Hypothesis Testing}
Today, it is considered more or less \emph{de rigueur} to accompany a proposed new method in computational linguistics with an \emph{automatic, formal, quantitative evaluation}. This reflects a healthy development towards greater objectivity in reporting results, but it also comes with a greater responsibility on the part of the researchers to ensure that the evaluation metrics provide a true measure of the accuracy of the proposed method. 

Because of the vast amount of digitized information now available to us, there is currently a unique possibility to develop and test methods for detecting language change. However, the amount of data limits the possibility to use expert help and manual efforts in the detection phase. It is also a limiting factor in the evaluation phase as there are to date no existing, open datasets for diachronic conceptual change that can be used for evaluation purposes. Specific to this problem is the grounding of diachronic conceptual change in a given corpus. When does a word appear for the first time with a new or changed sense in a given corpus? As a consequence, there are no automatic evaluation methods. Instead, there is a large variety of techniques, datasets and dimensions that are used in the existing literature. Most previous works have made use of manual evaluation. However, some have made use of WordNet for evaluation purposes. We argue that WordNet is not appropriate for evaluation for two main reasons. First, there is no indication in WordNet of \emph{when} a word's meaning changed or a new sense was added. Second, when datasets span hundred years or more, WordNet does not sufficiently cover the vocabulary or word senses in the dataset. The same holds for Wikipedia, which often covers changes but lacks time information \citep{holzmann2014insights}. In addition to the lack of data and resources for evaluation, there are no evaluation methods or metrics that have themselves been properly evaluated.

Note that downstream applications, \eg IR systems, can of course be evaluated in the normal way for such applications, which we will not describe here. Rather we will focus on methods for evaluating lexical change as uncovered by the methods surveyed here. A reasonable assumption would be that such an evaluation regime will also be useful -- at least in part -- for evaluating concrete downstream applications.

At least in the context of this literature survey, we would like to step back and see computational linguistics style formal evaluation as part of a larger endeavor, as a central and necessary, but not sufficient, component of \emph{(linguistic) hypothesis testing}. In particular, since the gold standard datasets which make up the backbone of our formal evaluation procedures are generally extremely expensive to create, there is an understandable tendency in our community to reuse existing gold standards as to the greatest possible extent, or even re-purpose datasets originally constructed with other aims in mind.\footnote{Or even generate synthetic, simulated data assumed to faithfully reflect authentic data in all relevant aspects.} However, such reuse may be in conflict with some assumption crucial to the original purpose of the dataset, which in turn could influence the results of the evaluation.

There are two central (typically tacit) methodological assumptions -- i.e. hypotheses -- made in the work described in the previous sections, and especially in work on diachronic conceptual change detection and classification (Section~\ref{sec:wsch}):

\begin{enumerate}
    \item \textbf{Applicability:} the proposed method is suitable for uncovering diachronic conceptual change.\label{hypo1}
    \item \textbf{Representativity:} the dataset on which the method is applied is suitable for uncovering diachronic conceptual change using this method.\label{hypo2}
\end{enumerate}

Since most current approaches are data-driven -- i.e. the data are an integral component of the method -- these two factors, while logically distinct, are heavily interdependent and almost impossible to keep apart in practice, and we will discuss them jointly here.

With a few notable exceptions, to which we will return below, there is also often a third tacit assumption:

\begin{enumerate}
    \setcounter{enumi}{2}
    \item \textbf{Falsifiability and control conditions:} positive evidence is sufficient to show \ref{hypo1} and \ref{hypo2}.\label{hypo3}
\end{enumerate}

Assumption \ref{hypo3} comes at least in part from the common practice of evaluating diachronic conceptual change using lists of attested such changes.

We will now take a closer look at these assumptions.

\subsection{Applicability and Representativity}
The first major difficulty when evaluating the results of diachronic conceptual change is the \emph{evaluation of a representation} $r \in R$ of a meaning of a word $w$ or a word sense $s_w$. When is $r$ a correct and complete representation of $w$ or $s_w$? Typically, this boils down to determining if a set of words, derived by clustering, topic modeling or the closest words in a word space, indeed correspond to the meaning of a word or word sense?  In the case of multi-sense tracking, it is also important that the set of representations in $R$ are a complete representation of $w$ such that all its senses are represented in correct way. The evaluation of individual word senses is analogous to the evaluation of word sense induction (see \citealp{Agirre:Semeval,navigli2012quick} for more details and an overview).  

Another related, more subtle, source of methodological muddles may be a misunderstanding of
what is being investigated. \citet{liberman-2013} points out that the
notion of ``word'' used in a paper by \citet{petersen-etal-2012} is
very far from how this term is understood by linguists, and the
purported statistical laws of vocabulary development as evidenced in
the Google Ngram dataset can be due to many other irrelevant factors,
foremost of which is varying OCR quality, but also ``tokenization'' as a faithful model of wordhood \citep{dridan-oepen-2012}.

Linguists have long recognized that ``language'' is a nebulous term, at best designating a convenient abstraction of a complex reality. This does not mean that any language sample should be considered equally representative, however. Especially corpus linguists have spent much intellectual effort on the question how to compile representative language samples, where it is clear that ``representative'' generally must be interpreted in relation to a specific research question. We mention this here, since we feel that it is important to be clear about what the changing entity is when we investigate lexical change. Given that linguists generally consider speech to be the primary mode of linguistic communication, are we happy investigating mainly written language, following a long tradition of ``written
language bias'' \citep{linell-2005} of general and perhaps especially computational
linguistics? Or given that the language should belong to every member of its speaker community, are we satisfied modeling the language of a select small social stratum \citep{henrich-etal-2010,sogaard-2016}? Whatever their answer to these questions are, authors ought at least to realize that they need to be addressed.

The second major difficulty concerns the \emph{comparison of word senses} (via their approximations) over time. Because the word senses are approximations derived from time sliced corpora, the representations at different time points can be different without there being any actual sense change. 
	Two factors can play a role: 
	\begin{itemize}
	\item[Fac. 1] Imagine a set of contexts $C$ that contain word $w$. If we split $C$ in two random sets $ C_1$  and $C_2$, such that $C_1 \cup C_2 = C$, the representations of $w$ will be different. Assuming that $|C_1|$ and $|C_2| \rightarrow \infty$ the difference in representation of $w$ for $C_1$  and $C_2$ should go to 0. However, this is rarely the case, our datasets are finite in size and we  see a difference in representations. Because we often use single genres of data, novels, news papers etc, we are likely to enhance this randomness effect; if a word is not used in a certain context due to missing underlying events, then the word sense will not be present. By using a mixed set of sources, we could reduce this effect. We see the same effect for representations of a word $w$ if $C_1$ and $C_2$ belong to two different time periods. 
	
	 Now, if $ C_1$  and $C_2$ derive from two adjacent time periods, the task of diachronic conceptual change becomes to recognize how much of the difference that is due to this randomness effect and how much is due to actual semantic drift. 
	 \end{itemize}
	 
	 \begin{itemize}
	\item[Fac. 2] Imagine that the representation of $w$ is a set of words $u_1, \ldots, u_n$ for time $t_i$ and $v_1,\ldots,v_n$ for time $t_j$. If each $v_j$ is a diachronic word replacement of $u_j$, then the entire representation of $w$ can be replaced between $t_i$ and $t_j$ without there being any change to the sense of $w$. While it is unlikely that all words are replaced between any $t_i$ and $t_j$, the risk of this effect increases the further apart the time periods. 
 	\end{itemize} 

In other words, in order to argue that some instance of lexical variation constitutes
a case of diachronic conceptual change based on (massive) corpus evidence, it it
generally not enough to ascertain that the variation correlates with
different time slices of the dataset. It is also necessary to ensure that no other relevant variables are different between the time
slices. The original Culturomics paper \citep{michel2011quantitative}
has been criticized for not doing this, by \citet{pechenick-etal-2015}
and \citet{koplenig-2017b}, among others. This is also held forth as a
strong point of the smaller COHA dataset by its creator
\citep{davies-2012}. This pitfall can be avoided by devising control
conditions, but even so the purported diachronic effect may
conceivably disappear for other reasons as well, \eg if some other
variable unintentionally correlates with time because of how the data
were compiled.

Another interesting reduction is the n-gram model, that automatically limits the amount of available information. To date, there has been little, if any, discussion in the diachronic conceptual change detection field to cover the effects of using n-grams rather than a full dataset with running text.\footnote{\citet[233]{gale-etal-1992} note that in their experiments on word-sense disambiguation, they ``have been able
to measure information at extremely large distances (10,000 words
away from the polysemous word in question), though obviously
most of the useful information appears relatively near the polysemous word (e.g., within the first 100 words or so).''} 
What happens when we remove words out of n-grams (which is the case when we only keep the K-most frequent words)? How many n-grams still have sufficient information left? What is the distribution of the remaining 1-, 2-, 3-, 4- and 5-grams after the filtering? This is particularly important when we consider those works that keep the K-most frequent words without normalizing over time, and hence have a modern bias among the kept words. If we start with equal samples over time, how many n-grams contribute over time?  

An important aspect of representativity is language coverage. While it is certainly true that the studies surveyed here are on a
much larger scale than any historical linguistic studies heretofore
conducted, it is nevertheless misleading to characterize
traditional historical linguistic investigations as ``based on small and
anecdotal datasets'' \citep[2]{dubossarsky-2018}. This ignores the
combined weight of the diversity of active observations painstakingly
and diligently made over two centuries on many languages and language
families by a large number of scholars highly trained in linguistic
analysis, observations which are continually shared and discussed in
the professional literature of the discipline.
Against this is set computational 
work on massive textual (published) datasets largely confined to one language
-- the norm -- or a typologically and geographically skewed 
sample of a few languages. While such work undoubtedly will contribute valuable data points to our collective knowledge of lexical change, in order to make solid linguistic claims about this kind of language change, it would be desirable to conduct equivalent experiments on as many languages as possible \citep[see \eg][]{bender-2011}.

\subsubsection{Factors Involved in Evaluation of Diachronic Conceptual Change Detection}\label{sec:evaltechWSE}

\paragraph{Granularity}
    
The first and most important factor that impacts evaluation is to determine the granularity on which to evaluate. Typically, change is evaluated with respect to change in the dominant sense of a word. That is, changes are not evaluated individually for all the senses of a word, instead, meaning change is evaluated for the form (text word or lemma), i.e. mixing all its senses. 
    Having a single representation per time period significantly reduces the complexity as it does not take into consideration what happens individually for each sense of a word. If a word has at most $s$ ($s\in S$) senses per time period over $t$ ($t \in T$) time periods, the number of unique senses is bound by $S\cdot |T|$. To compare all senses pair-wise between time periods there are at most $|T|\cdot$S$^2$ comparisons needed. If we wish to evaluate the \textit{similarity graph} created by the senses in each time period, where edges correspond to similarity between two senses $s_i \in t_i$ and $s_j \in t_j$, there are $S^|T|$ possible paths. In comparison, for the single representation case, the number of unique senses are $|T|$ and the number of necessary comparisons is $|T|-1$ 
    and there is only one path to evaluate. The number of time periods affect this complexity, and while some use yearly subcorpora, others use decades, reducing the time periods to compare by one order of magnitude.

\paragraph{Context}
What is considered the \textit{context} of a word differs largely between different works and is to some extent determined by the choice of dataset. A context ranges from 30 words surrounding $w$ \citep{sagi2009semantic} to the word before and after \citep{Gulordava}. When the Google N-gram data is used, the context can be at most a window of 5 words (from 4 words before or after, to the word $w$ being the first or last word, or 2 words before and after, the word $w$ being the 3rd word). What information is used as a context affects the representation. 

\paragraph{Words included in the evaluation}	
An important part of evaluation is to determine which words to evaluate; here two methods are employed; a set of pre-determined words, or the (ranked) output of the investigated method or methods.
	The former has the advantage of requiring less effort and reduces the need to conduct a new evaluation for each new run, with e.g., new parameters. The downside is, however, that the evaluation does not allow for new, previously unseen examples. 
	
	\begin{itemize}
		\item[] Pre-chosen testset
		\begin{itemize}
			\item positive examples (words known to have changed)
			\item negative examples (words known to be stable)
		\end{itemize}
	\item[] Output of algorithm
		\begin{itemize}
			\item on the basis of a pre-determined measure of change (e.g., largest/smallest cosine angle between two consecutive time periods)
			\item randomly chosen set of words
		\end{itemize}
	\end{itemize}
	
		Most commonly, single words are used in evaluation, but it is becoming increasingly common to study the relation between (known) word pairs. That means, two words, typically one that is under investigation and one that represents the changed word sense, are evaluated with respect to their similarity over time. If a change takes place between the pair, this is used to confirm the hypothesis of diachronic conceptual change. Examples include (\emph{gay, homosexual}) that become more similar over time, or (\emph{gay, happy}) that become less similar over time. Both would confirm the same hypothesis about change in meaning for the word \emph{gay}. 
	Thus far, word pairs have always been used in a pre-chosen fashion. Choosing the word pairs that have the highest amount of change increases the computations by a polynomial factor. If we assume that there are $n$ words at time $t$, and worst case, a new set of words for each time period, then there are $(n^2)^t$ pairs available. Typically, the situation would be much less extreme and only a fraction of the vocabulary is exchanged per time period (the more, the further apart the time periods are).  Moreover, the reference term to be chosen for judging the changes of a target term should itself have stable meaning over time. For example, when tracking the similarity between \emph{gay} and \emph{happy} in order to detect or understand the sense change of the former, one implicitly assumes that \emph{happy} does not undergo significant semantic change over the time period of comparison.

\paragraph{Evaluation technique} Evaluation can be conducted manually or automatically. The manual evaluation is done either with respect to intuition or pre-existing knowledge, or against one or more resources (dictionaries, encyclopedia etc). Automatic evaluation is performed with respect to external resources, \eg WordNet, or intrinsically where some evaluation metric is compared over time, \eg statistically significant difference in the direction of the word vectors. 

Evaluation of temporal analog search often follows IR style evaluation settings. For a given query a ranked list of analog terms is presented and the metrics like precision/recall \citep{tahmasebi2012neer} or precision@1, precision@5 and MRR \citep{zhang2016past} are used based on the rate of correct analogs found in the top ranks.
	
\paragraph{Change types included in the evaluation}	Evaluation for each word can be a binary decision; yes/no, there has been change, but it can also take the time dimension into consideration. The change is correct if it is found at the expected time point, or it is correct with a time delay that is measured. In addition to the binary decision, there are different change types, see Table \ref{tab:changetypes} for a list of change types considered in this literature. The more types are considered, the more complex the evaluation becomes. With one exception, different change types are considered only for sense-differentiated methods, while word level change groups all changes into one class. Typically, change means a shift in the dominant sense of a word. \Eg Apple becomes a tech company and adds a dominant meaning to the word Apple. However, its fruit sense is not gone but is very much valid.\footnote{Note, however, that in written standard texts this ``change'' will partly be an artifact of preprocessing; lower casing all text will increase the likelihood of conflating the common noun \emph{apple} and the proper noun \emph{Apple}.} Still, the change in dominant sense from `fruit/food' to `technology' is considered correct in a word level change setting. 
    
\begin{table}
 \centering
 \caption[Change types]{Change types investigated in the surveyed literature}\label{tab:changetypes}
 \begin{tabular}{ll}  
\toprule
Change type & Description  \\
\midrule
Novel word & a new word with a new sense, or for that word new sense\\
Novel word sense & a novel word sense that is attached to an existing word\\
Novel related ws & a novel word sense that is related to an existing sense.\\
Novel unrelated ws & a novel word sense that is unrelated to any existing sense.\\
Broadening & a word sense that is broader in meaning at a later time\\
Join & two word senses that exist individually and then join at a later time\\
Narrowing & a word sense that is broader in meaning at an earlier time\\
Split & a word sense that splits into two individual senses at a later time\\
Death & a word sense that is no longer used\\
Change & any significant change in sense that subsumes all previous categories\\
\bottomrule
\end{tabular}
\end{table}	

\paragraph{Time dimension}
	The time span of the data makes a difference in evaluation. The further back in time, the harder it is to evaluate since there are fewer resources that cover the data (e.g., no reference resources such as dictionaries/wordnets/wikipedias for historical senses, etc.) and fewer experts to perform in depth manual evaluation. The complexity is increased with the number of included time points. The more time points, the more complex the evaluation as there are more comparisons to evaluate. 
	
	The evaluation of time is an extremely complex matter; should it be done with respect to the outside world or the specific dataset under investigation? The complexity of the evaluation differs largely depending on the choice. To compare to the outside world means to make use of dictionaries and other knowledge sources to determine when a word came to existence, changed its meaning or added a sense. The resource or resources used for this determination need not be tied to the dataset used and there are regional varieties in uptake of new politics, technology, culture etc that in turn affect language use. 
    Newly coined terms, or senses can be due to an invention, one or a few influential sources, or an event and in such cases, be simpler to pinpoint in time. If the change, however, is due to a slow cultural shift or idiom that increases in popularity, it becomes very difficult to pinpoint the time of change. An analogy is that of fashion; when did the bob cut come into fashion? When the first ever person got such a haircut? Or the first celebrity showed it off on the red carpet (where is was better noticed and more likely to be duplicated)? Or when we can measure that a certain percentage of women had the hair cut as attested by \eg school pictures or driver's licenses. In manual attestation of diachronic conceptual change it is common to discuss the explanatory power of a sense in a given time, however, that is hard to translate into a specific time point. A more or less arbitrary threshold can be used to translate an increasing (or decreasing) curve into a binary yes or no that can be used to specify a time point. 
    
    If we wish to evaluate with respect to the dataset, there is an added difficulty compared to the above. If the word itself is not novel, then it requires word sense disambiguation to find the first occurrence of a new or changed sense; when was a word used in a specific sense for the first time in the dataset?  If existing sense repositories are not available, the senses must first be induced and then assigned to individual instances of a word in the dataset which is, to some extent, to solve half of the diachronic conceptual change problem. In addition, the results might be different for each dataset, and hence it is a time consuming procedure that must be repeated. However, disregarding differences between datasets might penalize certain datasets, and hence experiments, compared to others, \eg expecting an invention to appear in a dataset at invention time when in fact there might be a delay of  decades.  

For both methods there is a large difference between expecting to automatically find the first instance of change or expecting to find the change when it has gained enough momentum to be detectable by context-dependent methods.  An example of the differences in momentum but also the differences between datasets can be illustrated with the word \emph{computer}. An earlier common usage of this word was in reference to humans \citep{grier-2005}, but the `computing device' sense has been on the rise since the electro-mechanical analog computer was invented in the early 20th century and came to play an important role in the second world war, and its incidence has been increasing with the growing importance of digital computers. 
    The frequency of the word \emph{computer} in Google N-grams reaches over 0.0001\% in 1934 for the German portion, 1943 for the American English, and 1953 for the British English, meaning that a method evaluated on the latter dataset would be penalized by 20 years compared to one evaluated on a German dataset.\footnote{The word \emph{Rechner} was and is used in German as a synonym of \emph{Computer}.}    
 
Here we should also mention the sociolinguistic construct \emph{apparent time} \citep{mague-2006} and a similar idea which informs much work in corpus-based lexicography. Apparent time rests on the assumption that crucial aspects of our linguistic repertoire reach a stable state at an early age, say around the age of 20, so that \eg dialect studies can address diachronic development by recording age-stratified speaker samples synchronously, so that the language of a 70-year old is supposed to reflect -- in time capsule fashion -- current usage about 50 years ago. In a similar way, lexicographers assume that some genres are linguistically more conservative than others, and look for first appearances of new words or new word senses in news text rather than in fiction. Today, the intuition of dialectologists and lexicographers would conspire to single out social media texts as the main harbingers of lexical change \citep[\eg][]{fiser-ljubesic-2018}.

\subsection{Recommended evaluation procedure for diachronic conceptual change}
We recommend the following to be included in any evaluation procedure: 
\begin{enumerate}
\item \textit{Pre-chosen testset}: Compare the results for positive words, to other words from the same frequency bin, or to the average behaviour of all words, to reduce frequency bias. 
\item \textit{Grounding in the dataset}: Evaluate backwards referral to the original texts, e.g., by looking at randomly chosen N-grams or sentences, where the word under investigation occurs. 
\item \textit{Grounding in the outside world}: evaluate with respect to the outside world, e.g., dictionaries and encyclopedias. How well does the result correspond to the expected? In particular, if claims are made about language in general on the basis of results derived from the corpus. 
\item Consider conceptually and/or practically what happens if there is too little evidence in the text (for certain time periods) for a word: can meaning change be found?
\item Can different change types be differentiated in theory? In practice? This question should be answered even if the method is not used for differentiated change types in the study. 
\item Can the time of change be found?
\item How does the method scale up to more time points? This relates in particular to those that evaluate change on a few, far apart time points.
\item Always declare and give grounds for evaluation judgments: Yes, we consider this to be correct because $\ldots$, or No, we consider this instance to be incorrect because $\ldots$. 
\end{enumerate}

\subsection{Falsifiability and Control Conditions}\label{sec:falsifiability}

\citet{Dubossarsky-EMNLP17} highlight the importance of falsifiability, 
by devising a simple ``sanity
check'', creating control conditions where no change of meaning
would be expected to occur, and reproduce previous studies which have purported to
establish laws of semantic change, two proposed by
\citet{DiachronicWordEmb} and one proposed by themselves
\citep{Dubossarsky-2015}, finding that in the control
conditions, the reported sense change effects largely disappear or
become considerably smaller. They use the Google Books English fiction and sample 10 million 5-grams per year randomly from 1900--1999, each bin spanning a decade. Two control corpora are used, one randomly shuffles the 5-grams from all bins equally. The size of the vocabulary stays the same as in the original corpus, but most semantic change should be equally spread over the corpus, and hence not observable, or observable to a much lesser extent. A second control corpus is created by sampling 10 million 5-grams randomly from 1999, for 30 samples. Since all words are sampled from the same year, there should be no observable semantic change. Word representations are created using word counts, PPMI and SVD reduction of the PPMI matrix, and the three laws are evaluated on both the genuine corpus and the shuffled control corpus.  
All three laws were verified in the genuine corpus but also found again in the shuffled corpus. The three word representations were used with a cosine similarity measure on the second control corpus, the 30 samples drawn from 1999, and while the changed scores are all lower for the control corpus, they are significantly positive, showing that the proposed change measurements are affected by noise. Using analytic proofs, it is shown that the average cosine distance between a word's vectors from two different samples (using count-based representations) is negatively correlated with the word's frequency. 

The linguistic literature provides a wealth of fact and even more discussion about possible driving forces behind both linguistic
variation in general and linguistic change, typically accompanied by a
large number of empirical linguistic examples. As a minimal
methodological requirement, it would behoove authors proposing that a
computational method can bring new insight to the study of lexical
change in language, to demonstrate in a credible way that other kinds
of variation have been taken into account by \eg the experimental
setup, which crucially includes choice of appropriate positive \emph{and negative} data. Especially claims that seem to fly in the face of established truths in the field should be extremely carefully grounded in relevant linguistic scholarship. For instance, \citet{hills-adelman-2015} report a finding that semantically, the vocabulary of American English has developed in the direction of greater concreteness over the last 200 years, which seems to go against a proposed generalization about semantic change, namely that concrete vocabulary tends to be extended with abstract senses  \citep[383]{urban-2015}. A closer scrutiny of the methodology of the study reveals some questionable details. Thus, the list of crowdsourced concreteness ratings compiled by \citet{brysbaert-etal-2012} used in the study provides only one part of speech and one concreteness score per lemma, e.g. \emph{play} in this dataset is only a verb with a concreteness rating of 3.24 (on a $0-5$ scale). In a follow-up study \citet{snefjella-etal-2018} approach the same problem using a considerably more methodologically sophisticated and careful approach, but which still raises some questions. Building on work by \citet{hamilton-etal-2016}, they compute decadal concreteness scores for the COHA corpus (for the period 1850--2000) based on a small set of seed words assumed to have stayed stable in extreme concreteness and abstractness over the whole investigated time period, and find the same trend of increasing concreteness in the corpus over time. As an anecdotal indication of the accuracy of their approach they list the top 30 concrete and top 30 abstract (text) words that come out of their computation (\eg \emph{muddy, knives} vs. \emph{exists, doctrine}) and also report statistical correlations between the computed scores and several sets of human ratings, including those of \citet{brysbaert-etal-2012}. However, looking at the scatterplots provided by  \citet[6]{snefjella-etal-2018}, it is clear that the computed scores inflate concreteness compared to the human ratings, and in particular at the more abstract end of the concreteness range.\footnote{This does not in itself invalidate their result, of course. If this tendency is consistent over time, we are still seeing a diachronic increase in concreteness of the same magnitude that they report.} Further, if we POS tag the results\footnote{Their resulting data are available in their entirety at \url{http://kupermanreadlab.mcmaster.ca/kupermanreadlab/downloads/concreteness-scores.zip}} we note that many function words (\eg determiners and prepositions) come out as highly concrete (\eg \emph{the} is very close to \emph{muddy} for some of the decades), whereas they cluster consistently at the abstract end in the human ratings. The results reported by \citet{hills-adelman-2015} and \citet{snefjella-etal-2018} are very interesting to a historical linguist and deserve further study, but their studies should be replicated, with clear control conditions informed by awareness of historical linguistic facts, before any secure conclusions can be drawn.

\subsection{Datasets and Testsets}\label{Evaluation_diach_replacement}
We briefly discuss here some of the datasets that have been used in the literature (or that could be used) for automatic evaluation of lexical and semantic change over time. The list is not exhaustive. 

Large scale data is the basis for effective and reliable word analysis. The largest available historical corpora\footnote{We follow general practice in the computational linguistic literature and use the term ``corpus'' even about datasets such as the Google Books Ngram dataset, which are not strictly speaking corpora in the corpus linguistic sense, in other words are not composed of complete texts.} with a comprehensive representation of several languages in the past are probably the \textbf{Google Books Ngram datasets}, which are available online.\footnote{\url{http://storage.googleapis.com/books/ngrams/books/datasetsv2.html}} The English datasets, for example, cover about 5\% of all books ever published. The total size of Ngram lists ($n={1,2,3,4,5}$) is quite large (about 0.3 trillion words), which typically necessitates provision of effective and scalable methods. The data are organized as ngram counts by every year and are available for several languages including English, French, Spanish. There are also some specialized English corpora available, such as American English, British English, and English Fiction. Google Books Ngram datasets come in two versions: version 1 (compiled on July 2009) and version 2 (compiled on July 2012).
Due to the scarcity of available books, the data for the period before the 19th century tends to be sparse (e.g., it is estimated that only about 500,000 books were published in English before the 19th century). In addition, there is the potential for a high number of OCR errors for the texts written in earlier centuries due to deterioration in the quality of paper, non-standard fonts, many spelling variants, and so on. According to the dataset creators, the rate of OCR errors is however limited, as ngrams that appear less than 40 times across the corpus have been removed. Independent analysis \cite{pechenick-etal-2015} has highlighted a number of limitations of the datasets. 
One concern was that they are increasingly dominated by scientific publications rather than by popular works. Nevertheless, Google Books Ngrams have been used frequently for not only diachronic linguistics but also for culturomics studies aiming to capture societal and cultural changes over time \cite{michel2011quantitative}.

Another corpus commonly used for diachronic conceptual change analysis is the \textbf{Corpus of Historical American English} (COHA) developed by Brigham Young University (BYU) \citep{davies-2012}.\footnote{\url{https://corpus.byu.edu/coha/}} Unlike the Google Books Ngrams datasets, COHA has been compiled with decade-level granularity. An important advantage of COHA is that it is based on a stable distribution of diverse document genres for each decade. The document genres (\eg fiction, magazine, newspapers) were chosen according to the Library of Congress categorization scheme.\footnote{\url{http://www.loc.gov/catdir/cpso/lcco/}} COHA contains over 400M words given as ngrams ($n={1,2,3,4}$) collected from about 107K documents published from the 1810s to the 2010s. Each ngram is associated with its count in every decade. Part-of-speech tags are also available. According to COHA creators, the corpus is 99.85\% accurate,
which means, on average, there is one error for about every 500--1000 words. COHA is also available as a conventional full-text corpus. \footnote{\url{https://www.corpusdata.org/}}

The two above-mentioned datasets are substantially different. The Google Books Ngrams datasets have been compiled using all the books digitized through the
Google Books initiative. In contrast, COHA contains carefully selected texts and it is characterized by a stable rate of different document genres across decades. However, many rare words are not present in COHA making it useful only for the analysis of relatively common terms. On the other hand, the Google Books Ngrams datasets are more effective in this regard, given their large size. 
The weakness of both datasets is relatively short text window used for computing ngrams (up to 5 and 4 words), which makes it impossible to capture long distance word dependencies as also noted by \citet{tang-2018}, unless one uses the full texts. Finally, pre-trained, downloadable word embeddings for historical text are available \cite{DiachronicWordEmb} based on the Google Books Ngrams and partially on COHA.\footnote{\url{https://nlp.stanford.edu/projects/histwords/}}

The \textbf{Corpus of Contemporary American English} (COCA) \cite{Davies2010} is another balanced corpus from BYU.\footnote{\url{https://corpus.byu.edu/coca/}} As it spans a much shorter time frame (1990--2017), it could be used for analyzing short-term and recent changes in word meaning. The dataset comprises more than 560 million words, and 20 million words per year. COCA has been compiled from spoken, fiction, popular magazines, newspapers and academic texts.

\textbf{DiAchronic TExt corpus} (\textbf{DATE}) covers the years between 1700 to 2010, combining data from the COHA corpus, the training data from SemEval-2015 Task 7: Diachronic Text Evaluation \cite{popescu2015semeval} and the portion of \textbf{The Corpus of Late Modern English Texts, version 3.0}\footnote{\url{https://perswww.kuleuven.be/~u0044428/clmet3_0.htm}}, (\textbf{CLMET3.0}) \cite{diller2011european}. 
CLMET3.0 is a collection of public domain texts obtained from various online archiving projects; it contains about 34 million words and covers the period from 1710 to 1920, divided into three periods.

The \textbf{TIME corpus}\footnote{\url{https://corpus.byu.edu/time/}} was generated from new articles published in a single source: TIME magazine from 1923 to 2006.
The 275,000 articles processed provided 100 million words of text. Other similar collections of a single source are: the \textbf{Times Digital Archive} \footnote{\url{https://www.gale.com/uk/c/the-times-digital-archive
}} (from 1785 to 2012) and the \textbf{New York Times Annotated Corpus} \footnote{\url{https://catalog.ldc.upenn.edu/ldc2008t19}} (from 1987 to 2007). 

Among the other corpora\footnote{More examples of available corpora are also listed in: \url{http://davies-linguistics.byu.edu/personal/histengcorp.htm}} worth noting are the \textbf{Lampeter Corpus of Early Modern English Tracts},\footnote{\url{http://clu.uni.no/icame/manuals/LAMPETER/LAMPHOME.HTM}} the \textbf{Corpus of Early English Correspondence} (\textbf{CEEC}),\footnote{\url{http://clu.uni.no/icame/manuals/CEECS/INDEX.HTM}} the \textbf{Diachronic Part of the Helsinki Corpus of English Texts},\footnote{\url{http://clu.uni.no/icame/manuals/HC/INDEX.HTM}} and the \textbf{Diachronic Corpus of Present-Day Spoken English} (\textbf{DCPSE}).\footnote{\url{http://www.ucl.ac.uk/english-usage/projects/dcpse/}}

\textbf{Project Gutenberg}\footnote{\url{http://www.gutenberg.org/}}, one of the oldest digital libraries, was founded in 1971 and provides over 50k digitized books from the public domain in plain text and other formats. 
 \textbf{HathiTrust}\footnote{\url{http://www.hathitrust.org}} is a large-scale collaborative repository containing digital content from research libraries and other sources in collaboration with Google Books and Internet Archive projects. The library offers over 13 million books via full text search. \textbf{Early English Books Online (EEBO)}\footnote{\url{https://corpus.byu.edu/eebo/}} is a collection of texts created by the Text Creation Partnership, which contains 755 million words in 25,368 texts from the 1470s to the 1690s. \textbf{CLARIN ERIC}, a large European research infrastructure based on language technology and language resources, has published a list of 70 historical corpora. It covers a large number of (primarily European) languages.\footnote{\url{https://www.clarin.eu/resource-families/historical-corpora\#list-of-publications-on-historical-corpora}}
The \textbf{Internet Archive}'s text collection\footnote{\url{https://archive.org/details/texts}} contains more than 15 million freely downloadable digitized texts in different languages (mainly in English) amassed from various libraries and cultural heritage institutions. 

Court rulings, decisions, and criminal trials make up a genre of resources that has survived in recognizable form through centuries. For example, \textbf{Corpus of US Supreme Court Opinions}\footnote{\url{https://corpus.byu.edu/scotus/}} contains around 130 million words from 32k decisions of USA Supreme Court dating from the 1790s until the present. 
The \textbf{Proceedings of the Old Bailey}\footnote{\url{https://www.oldbaileyonline.org/}} (1674 -- 1913) are a fully searchable body of the records of close to 200,000 criminal trials at London's central criminal court providing a unique glimpse into the lives of non-elite people in London over a relatively long time. 

Other resources can be more domain-specific. 
Such datasets can be used to track sense changes of terms in specialized domains.
For example, \textbf{Hansard}\footnote{\url{https://www.hansard-corpus.org/}} is a specialized corpus containing almost every speech delivered in the British Parliament from 1803 to 2005 (over 7.5 million texts). Besides the access to the full texts,\footnote{\url{http://www.hansard-archive.parliament.uk/}} users may also search within the corpus.\footnote{\url{https://www.hansard-corpus.org/x.asp}}
 \textbf{Medline}\footnote{\url{https://www.ncbi.nlm.nih.gov/pubmed/}} is a repository containing articles in medicine and related fields, published from 1950s on. The \textbf{Association of Computational Linguistics (ACL) Anthology} corpus \cite{radev2013acl}, which spans the period from 1965 to 2013, contains scientific papers published in ACL venues. \textbf{Amazon Product Reviews}\footnote{\url{https://snap.stanford.edu/data/web-Movies.html}} and \textbf{Amazon Movie Reviews}\footnote{\url{https://snap.stanford.edu/data/web-Movies.html}} are other examples of specialized datasets; they cover 18 and 10 years, respectively.

Thesauri are other resources that can support diachronic analysis over time. One example is the \textbf{Historical Thesaurus of English}.\footnote{\url{http://historicalthesaurus.arts.gla.ac.uk/guide/}} It contains nearly all words that were recorded from Old English period until the present. The thesaurus was created based on the Oxford English Dictionary (OED) data as well as on the second edition of the Thesaurus of Old English. It contains the records of word use with the date information as well as style, frequency, or geographical labels. Word synonyms are arranged in chronological order based on their first recorded date. As words' last recorded dates are also sometimes available, this resource can be used for finding obsolete words or obsolete meanings of presently used words. \textbf{Roget's Thesaurus} \citep{roget-1852} is another widely used English-language thesaurus, which has appeared in numerous editions and variants since its first appearance in the mid-19th century. Its past editions offer unique opportunities to explore lexical information as recorded by researchers in the past.

We note that, to the best of our knowledge, there is no ready lexical reference system for historical language that would be similar to WordNet \cite{miller1990introduction} in which data on historical senses of words are represented in a structured format using synonym sets and their relations.

\subsubsection{Testsets for lexical replacement}
When it comes to finding diachronic replacements or estimating the across-time similarity, any document collections with a sufficiently long time span and large enough size can be used as underlying datasets. For example, the \textbf{New York Times Annotated Corpus} (NYT) and the \textbf{Times Digital Archive} have been used for these purposes. 

Test sets for both of these datasets have been created for selected time periods \cite{tahmasebi2012neer, zhang2016past}. \citet{tahmasebi2012neer} introduced a test set\footnote{\url{http://tahmasebi.se/project/neer/}} composed of 86 co-references that correspond to 33 distinct entities that had changed their names over time (\eg \emph{Pope Benedict XVI}, \emph{Cardinal Joseph Ratzinger}). All the terms exist in NYT Corpus and have at least 5 occurrences during the year when the name change occurred. The authors provided also an extended version of the test set containing named entity pairs whose name change happened outside of the time span of the NYT corpus or that occur just a few times in the text.

The short-term test set \footnote{\url{http://www.dl.kuis.kyoto-u.ac.jp/~adam/temporalsearch_short.txt}} for NYT that was used by \citet{zhang2016past}, contains 52 pairs of term analogs. The terms were selected from three equal size time periods: [2002,2007], [1992,1996], and [1987,1991]. The authors have also provided an extended version\footnote{\url{http://www.dl.kuis.kyoto-u.ac.jp/~adam/temporalsearch_short_extended.txt}} \cite{Zhang:2017:TAR:3132847.3132917} which in total has 225 term pairs for the two time periods compared [2002,2007] and [1987,1991] and 100 term pairs for [2002,2007] and [1992,1996]. Three types of entities are represented: persons, locations and objects. Persons include presidents,
prime ministers and chancellors of the most developed and populous countries (\eg USA, UK, France) as well as the names of
popes and FIFA presidents. Locations include names of countries or cities (\eg \emph{Czechoslovakia}, \emph{Berlin}) that have changed their names over time, split into several countries, merged into another political system, or became new capitals. Finally, objects contain terms denoting devices (\eg \emph{iPod}, \emph{mobile phone}, \emph{dvd}), societal phenomena (\eg \emph{email}), companies/institutions (\eg \emph{NATO}, \emph{Boeing}) or other objects (\eg \emph{letter}, \emph{euro}).
The authors also provided a long-term test set\footnote{\url{http://www.dl.kuis.kyoto-u.ac.jp/~adam/temporalsearch_long.txt
}} for the Times Digital Archive, which is composed of 400 query pairs collected from diverse time periods covering the 20th century. Due to sparse data in the past, the time periods in the more distant past are longer than the ones in near past so that all contain data of more or less similar size, which is needed for effective word embedding training.

\subsubsection{Data Quality}

Determination of word senses and their changes must be done on high quality datasets that are representative of the language used at particular time. We briefly discuss here some of the issues related to the choice of datasets.

There is an over representation of English datasets in literature. This is a result of easily available datasets, with sufficient volume and time span. However, the current bias towards English, indicates the need for evaluating and developing methods for other languages as well. Not all languages have such volumes of digitized historical texts, which means that despite the shift towards methods for tackling large scale data, we need to, simultaneously, keep investigating methods that can detect change from small scale or fragmentary data.
Moreover, few existing datsets are of a scale large enough to permit investigation of less common words. This is especially critical for earlier periods for which the data tends to be sparse and subject to more OCR errors.

The representativeness of datasets is another issue that determines if one can accurately represent the language used at a particular time. Naturally, due to the way they are created, many corpora tend to reflect written language, which may differ significantly from spoken language (especially, considering that the skills of writing and reading were often the privilege of narrow groups in the past). Another concern is the need to balance the different document genres over time as well as the geographical distribution from where the content is collected. In general, good dataset representativeness means maintaining, as much as possible, realistic balance over many text attributes including document genre, topic, originating location, author's age, gender, social position, among others. As preparing representative datasets is often difficult, any obtained results should at least be compared over two or more corpora, especially ones that have different or complementary
characteristics.

We note that the results derived from a single dataset without full coverage of a language, provide a picture of the language use in that particular dataset. Such insight is an excellent resource for archival search and help in interpreting the content of that particular dataset. While results of this kind do not offer a complete view of what has happened in a language, they do offer a complementary view. Results derived from, for example, a dataset containing texts created solely by religious institutions might not correspond to other society segments (e.g., peasants, merchants, or even females), but will offer us a view of a historical variant of a language, and can possibly offer insights into the general language as well. However, important when presenting results is to reason about representativeness of the data, and to not overstate the results.

\section{Applications and Online Systems}\label{sec:appl}
\subsection{Systems Supporting Analysis}

Several online visualization systems and demonstrations supporting manual analysis have been proposed to complement the research methods for automatic diachronic conceptual change detection. They allow for verification of the results obtained from automatic methods or provide novel means for manually determining the diachronic conceptual change and its character. In such systems, the level of interactivity and user freedom in querying the data as well as the possibility for multi-dimensional analysis play important roles. In addition, the usability criteria are an important and standard evaluation focus of user interfaces. Visualization systems tend to be attractive not only to professionals and scientists, providing either a complement to automatic analysis or serving as the main tool for the analysis. They are also particularly suited for use by lay users especially if the systems are intuitive and highly usable.

The recent enhancement of automatic definitions generated by the Google search engine for ``definition queries'' is one example of a popular way to disseminate information on word origin and changes in popularity. For an input word, the standard word definition is complemented with a brief description of the word's origin as well as its frequency plot over time. Although users can see the count of a word over time, they essentially have to reason about word meaning change over time on their own.

The Google Books Ngram Viewer\footnote{\url{https://books.google.com/ngrams}} is a popular online application for observing and analyzing the frequency of a word or ngram over time; it is based on Google Books Ngrams datasets. It has been frequently used for digital humanities research \citep[\eg][]{michel2011quantitative,acerbi-etal-2013,bentley-etal-2014,pechenick-etal-2015,iliev-etal-2016}. The temporal frequency plots of several words or ngrams can be contrasted with each other. Users can choose wildcard searches (by putting * in place of a word in given phrase to obtain the top ten substitutions) or do a case insensitive search. Searches based on POS tags (e.g., plotting frequencies of \emph{tackle} separately as verb and separately as noun) are also possible as is frequency plotting based on five composition operators (e.g., summing or subtracting the frequencies of several expressions). Furthermore, inflection-oriented searches can be done (e.g., searching with \texttt{book\_INF a hotel} returns results for \emph{book, booked, books}, and \emph{booking}). The Ngram Viewer allows the identification of ngrams at the start and end of sentences and adjusting plots to consider ngrams only at either of these positions. 
The Ngram Viewer provides dependency relations by the => operator; for instance, to understand how often \emph{tasty} was used to modify the word \emph{dessert} (\texttt{tasty} => \texttt{dessert}). This combines frequencies of all instances in which the word \emph{tasty} modifies \emph{dessert} including \emph{tasty frozen dessert, tasty yet expensive dessert}. Dependencies can be further combined with wildcards (e.g., \texttt{drink=>*\_NOUN} to track frequencies of expressions containing different kinds of beverages as nouns). 
Nevertheless, because the viewer is based on frequency signals of words (i.e., probabilities of seeing a given ngram, or of a set or composite of ngrams in each given year), it does not provide a means for immediately portraying exactly how a term was used in the past or when its meaning transitions occurred. The viewer is more suited for culturomics or cultural text mining studies similar to other tools for general-purpose interactive exploration of diachronic corpora \citep[\eg][]{michel2011quantitative,odijk2012time,eijnatten2014using,jatowt2016historycomparator}.

The online interfaces to various corpora created by Mark Davies at Brigham Young University\footnote{\url{https://corpus.byu.edu/}} provide powerful options for users, without the need to write any code. For example, they provide frequency plots over time, comparisons across selected decades, examples of keywords in context (KWIC) at different time points, and listings of collocates. 

\citet{hilpert2008assessing} applied a variant of hierarchical clustering called variability-based neighbour clustering. The idea is to cluster adjacent time units (hence neighbour clustering) if the frequency of target term does not change much. The resulting dendrogram allows for identification of time points of large frequency change, which may indicate possibility of diachronic sense shifts. No context is used for a target word because the method relies only on frequency information of a query word, which limits the applicability of this method in visualizing diachronic conceptual changes of words.

\citet{rohrdantz2011towards} used LDA topics as representation of words' different senses and tracked their intensity change over time in a similar fashion to \citet{sagi2009semantic} who applied LSA for similar purpose. Twenty-five words before and twenty-five words after the target word were used as the context of the term, following suggestion given by \citet{schutze-1998} for automatic sense disambiguation. This allows for noticing various kinds of semantic changes of words such as the broadening or narrowing of senses and first occurrences of senses, especially as all the topics are showed over time in a single view. According to the authors, an interactive visualization approach provides the possibility to
detect key patterns at-a-glance, and, at the same time, to observe the details of the data by zooming in on the occurrences of particular words in their contexts. Additionally, pairwise comparisons of word senses with respect to their shared contexts were also displayed. The authors, however, restricted their system to only a short time period, demonstrating results on New York Times Annotated corpus. 

\citet{heylen2012looking} proposed using Multidimensional Scaling \cite{cox2008multidimensional} technique with a window length of 4 words before and after the target word and Pointwise Mutual Information for weighting context terms. 
They took this approach because they had observed that earlier automatic approaches that use distributional models use them in an indirect, black-box fashion, failing to indicate particular semantic properties and relations that play key roles. Motion charts from Google Chart tools were then used for visualizing occurrences of nouns in a 2D representation of their semantic distances. Hovering a mouse pointer over the bubbles denoting nouns shows text in which each noun occurs to let users interpret the precise meaning of the occurrence of the noun. The authors focused in their case study on Dutch words extracted from Dutch newspaper articles published from 1999 to 2005, which were organized in 218 synsets containing 476 nouns in total. Although they did not use the motion feature of the charts, they admitted to the possibility of tracking over time the centroid of tokens of a target word in the semantic space and showing the dispersion of the tokens around the centroid.

\citet{odijk2012time} demonstrated the interactive environment that visualizes information on volumes and correlations of words and documents across time. Similar to \citet{michel2011quantitative}, their focus leans more towards understanding historical and social aspects rather than on the shifts in word semantics. 

\citet{hilpert2015meaning} utilized animations in the form of animated scatterplots to portray changes in patterns over time as a metaphor of a petri dish.   
The authors have focused on a single pattern ``many a [noun]'' as a case study. Spots in the graphs represent nouns involved in the same pattern and are plotted next to each other if they have high similarity. Their size is bound to frequency of a noun or noun type in a particular time unit. During the animation the changes in the size and distances of spots provides knowledge of different uses of the pattern over time.

Dimensionality reduction techniques such as PCA, LSA or the quite popular t-distributed Stochastic Neighbor Embedding (t-SNE)\footnote{\url{https://lvdmaaten.github.io/tsne/}} \cite{maaten2008visualizing} started to be used for plotting ``trajectories'' of word meaning over time in vector spaces through 2D plots. By showing points that represent the meaning of the same words at different years or decades on the same 2D plot \citep[see, \eg][]{hamilton-cultShift,kulkarni2015statistically} and optionally connecting them with arrows, one static view shows how the words changed their semantics, by following their ``trajectories''. Typically some background reference terms are added along such ``trajectories'' to ground and explain the meaning.

\citet{theron2015diachronic} demonstrated an interactive visual tool for advanced analysis of the data of Spanish historical dictionaries. Their approach is unique as they utilized different editions of Spanish language dictionaries over time: the 1780, 1817, 1884, 1925, 1992 and the 2001 editions provided by the Royal Spanish Academy. 
In this method the dictionary editions are arranged in a matrix in columns (right to left in chronological order), while the meanings of a word are placed on the rows (top to bottom in ascending order). Lines are drawn to connect the related meanings across time, where the connection is computed using NIST or BLEU metric \cite{zhang2004interpreting}, which are frequently utilized in evaluating machine translation or summarization accuracy. Starting from the most recent dictionary, a particular meaning is connected to its closest meaning in the previous dictionary; if there is nothing that satisfies the predefined similarity threshold then the procedure is repeated for the older dictionary. Connecting lines can have branches in cases of bifurcation or merging of meanings. The authors call the resulting diagrams diachronlex diagrams. Diagrams can be further improved by collapsing nearby lines having similar temporal patterns or by simplifying branches. Furthermore, users with editing rights can annotate meanings or change their associations.

\citet{martinez2016design} introduced a system called \emph{ShiCo} for visualizing shifting concepts of Dutch words over time. It measures the changes in words used to refer to concepts based on a model previously introduced in \citet{kenter2015ad}. The model used requires a series of semantic spaces which are constructed by training word embeddings (e.g., word2vec) for different units of time (typically, each unit spans 10 years) and is based on two steps, generation and
aggregation. The generation step works in an iterative fashion such that an initial seed set is taken which typically consists of a small number of user-provided terms. Then words most semantically similar to the seed set are found based on the computed similarity values between word embeddings. A semantic graph is constructed from these terms and central terms are extracted using graph centrality measures. Next, the central terms are used as the seed set for the next iteration of the generation step. In the
aggregation step, the lists of words produced in the generation step are aggregated to produce the final word lists to be presented to the user.

The visualization is composed of two kinds of complementary graphs: a stream graph and a series of network graphs. The former shows color-differentiated streams for each term; the stream sizes represent the relative importance of the term in period. The importance is measured as either a term count in each time unit or as a sum of the similarities of the term to the seed terms. The network graphs for each time unit display the relations between terms in this time unit.

\citet{xu2017temporal} used word clouds and a heatmap for visualization of term semantics shifts by utilizing sequentially trained embedding vectors with initialization based on previous time periods, as proposed in \cite{TempAnal-NeuralLangModel}. They used the New York Times and National Geographic magazine articles as underlying datasets, the latter spanning about 110 years. As in \cite{martinez2016design}, Temporal Semantic Similarity Word Cloud is used to show, for a given time unit, terms most similar to a target query. As in standard term clouds, the font size of words is bound to their similarity value to the query word. Heatmap views let users see the similarity values of the terms most similar to the target term in each year, using colors. The y-axis of the heatmap is a list of words and the x-axis is a list of temporal periods such that for each given word (each row) one can understand the pattern of the changes in the similarity of this term to the target term (also called anchor term). The results from the two different datasets were contrasted with each other. 

For visual support in an analysis of lexical replacement, \citet{Mazeika} focused on semantically similar entities from different time periods. They extracted named entities from the YAGO ontology and provided a visual analytics tool to analyze the evolution of named entities of the New York Times Annotated Corpus. No name changes were tracked but the tool offers a visualization of the evolution of an entity in the relation to other entities.

\citet{Jatowt:2014:FAS:2740769.2740809} described an analytical framework that incorporates across-time self-similarity plotting as well as a decade-to-decade similarity heatmap, across-time sentiment analysis, diachronic comparative word analysis and key context term listing, using both Google Books Ngrams and COHA datasets. The signals from the different analyses were proposed to be combined to allow for reasoning on diachronic conceptual change. Two different word representations were used, a simple bag-of-words and distance-aware bag-of-words, where the distance is measured as a relative position of a context term from the target term. For simplicity, the sentiment values of context words were assumed to remain stable over time when computing the change of the sentiment of the target term's context. 
Their work resulted in the development of online investigation system for diachronic conceptual change analysis \cite{jatowt2018}.\footnote{\url{http://tinyurl.com/WordEvolutionStudy}} The system allows for detailed analysis of diachronic conceptual changes from diverse viewpoints based on word context comparison across-time, temporal term cloud and term tree generation as well as for contrastive analysis of word pairs of larger groups of words such as synonyms. The results can be investigated on both Google Books Ngrams and COHA datasets, using Pearson correlation, cosine similarity and Jaccard similarity used as similarity measures of word representations from different time points. Diachronic conceptual change over time is contrasted with term frequency plots (as was also suggested in \cite{TempAnal-NeuralLangModel}) since both provide a more informed view on how often and in what sense a term was used in the past. Moreover, any conclusions drawn from semantic change plots should be taken with caution when a target term had low utilization rate as demonstrated in its frequency plot. Also, the degree of the target word's change over time is displayed in reference to the average change of words in the same frequency bin as the target word.
The system has a novel feature that allows for investigation of changes in individual context terms over time in the form of a time-enhanced term cloud and time-enhanced term tree. This framework provides the unique functionality to track the semantic shifts of entire concepts represented as word sets (e.g., the concept of a \texttt{vehicle} represented by words like \emph{auto}, \emph{automobile}, \emph{car}, \emph{truck} and so on.
Another interesting option of the proposed system is the possibility to visualize only those context terms that precede or follow the target word in sentences and the option to contrast context terms of two compared terms as joint, color-coded, time-enhanced term clouds.

In general, interactive visualization and analysis of diachronic word change belongs to an emerging and powerful research field of interactive visualization for computational linguistics \cite{collins2008interactive}. Its purpose is to let users understand models of language and their abstract representations, and to uncover patterns in language through visual means. In view of the inherent complexity of tracking word senses and understanding their shifts over time, we expect an increase in the availability and popularity of visual, interactive approaches to diachronic corpora. Furthermore, it is often impossible to precisely determine the exact time of sense change, let alone accurately determine the nature of the change. Hence, several conflicting views or hypotheses can be simultaneously valid, causing scientists and professionals to only use automatic approaches to objectively support final judgments. Novel approaches for visualizing the histories of words from diverse aspects would be helpful to facilitate this process. A similar conclusion were reached in \citet{tang-2018}, which listed further exploration of data visualization techniques for hypothesis justification as one of the core issues to be solved.

\subsection{Other Applications}
As mentioned, the findings observable by methods for language change analysis and related models of diachronic conceptual shifts can be incorporated or taken into account into variety of applications that deal with texts. This is especially the case with texts written in the past or ones belonging to long-term document archives. Some promising directions in these areas are discussed below. 

\textbf{Culturomics}. \citet{michel2011quantitative} 
demonstrated changes in the frequencies of selected words that indicate higher level cultural or abstract changes occurring in society. As one example, they contrasted the popularity plots of the words \emph{men} and \emph{women} to provide evidence for the increasing social role and emancipation of women in recent decades. The study by \citet{michel2011quantitative} inspired similar studies which used both the Google Books Ngrams datasets \citep[\eg][]{acerbi-etal-2013,bentley-etal-2014,pechenick-etal-2015,iliev-etal-2016}, and other diachronic corpora \citep[\eg][]{hills-adelman-2015,snefjella-etal-2018}, as well as other languages \citep[\eg][]{viklund-borin-2016}.

While the implicit approach behind Culturomics relies on investigating changes in the usage intensity of words as well as data on their first appearances, its extension should also consider fluctuations in the meaning that words represent. \citet{tahmasebi2017uses} demonstrated the utility of automatic sense detection for archive users and for digital humanities research. They proposed a sense-based approach to capture changes related to the usage and culture of a word.

\textbf{Document similarity computation and information retrieval}. \citet{morsy2016accounting} proposed using the information about language change in document similarity computation. \citet{BerberichBSW09,holzmann2012fokas} discussed direct application of methods for detecting semantic shifts of words over time to IR, mainly for query suggestion. We believe that closer link between the approaches discussed in this paper and the components of search engines as well as information retrieval techniques in general, should be further pursued for improving information access in document archives.

\textbf{Diachronic text evaluation}. Diachronic conceptual change detection and characterization can be also applied for the purpose of document publication (origin date) detection \cite{Frerman2016}, also called \emph{Diachronic Text Evaluation} (DTE) \cite{popescu2015semeval}. Most of the state-of-the-art solutions for DTE rely on the information about word occurrence in the past with the underlying hypothesis that if \emph{a document contains many words that were common at given time, it is likely that the document was created/published at that time}. This is especially the case if the words were rarely used in other periods \citep[see, \eg][]{kanhabua2009using, chambers2012labeling,szymanski2015ucd,jatowt2017}. Including information on the diachronic meaning change can further improve the performance of diachronic text evaluation as studied by \citet{Frerman2016}.

\textbf{OCR error correction}. Automatically detecting and correcting errors in OCR-processed historical texts \cite{chiron2017icdar2017} could also benefit from the research on diachronic conceptual change in a fashion similar to the document creation date detection task.

\textbf{Comprehensibility of past texts}. Modules for estimating reading difficulty could be incorporated into archival search engines and recommendation systems, so that past texts which current users could unerstand are provided. Many of the methods described in this paper can be then of help here, because changes in word meaning over time can reduce the ease of reading and comprehension, as suggested in \cite{tahmasebi2013role}. In addition to extending readability indexes, this methods might be used to highlight words in text that have undergone considerable change and/or for clarifying their past senses to improve comprehension by average readers. The latter objective, though not exactly from the viewpoint of diachronic conceptual change, was approached by \citet{Tran:2015:BPS:2684822.2685315} who proposed re-contextualizing past texts by enriching them with context extracted from Wikipedia.

\textbf{Studies in domain-specific collections}. Detecting shifts in word senses can be limited to specific domains such as scientific papers and/or to specific types of terms. For example, \citet{degaetano2017diachronic} analyzed differences in frequency, meanings and underlying temporal scopes of temporal expressions used in scientific writing from 1665 to 2007. 

\textbf{Spatio-temporal word sense variation}. In addition to word senses changing over time they can vary along other dimensions orthogonal to time, for example, space. Combined spatio-temporal sense tracking can benefit from computational approaches developed for temporal sense analysis.

\textbf{Recommending Words for Analysis}. Etymological knowledge is not only interesting for professionals who work with texts like linguists, historians, or librarians, but also to the wider public. For example, many books on word histories are published for wider readership, or Google search engine enhances the definitions of words with brief summaries of their etymologies, as well as their frequency plots over time, for definition-like queries. 
We think that computational approaches and especially online interactive systems could help to further disseminate knowledge of word etymologies. For this, it would be important to recommend interesting words to be exploreed by non-professional users. 
Past meanings of words like \emph{gay} or \emph{nice}, for example, tend to surprise average, lay users. However, most online systems for exploring word senses and diachronic conceptual change require users to provide words as the input without recommending any of them. Since users may not know what to search for, recommending sample queries to explore and learn about could be a useful option to attract users. One possible solution could be to recommend unique or specific input words based on the shapes of their self-similarity plots over time \cite{jatowt2018} (e.g., words that retained stable senses over long time or that underwent significant semantic shifts within short time frames). 

\textbf{Semantic change in social media and the web}. While we have mainly discussed approaches for long-term sense tracking and analysis in this paper, recently researchers have also focused on diachronic change over shorter time spans such as several years \cite{dodds-etal-2011,danescu2013no,eisenstein-etal-2014,goel2016social,deltredici-fernandez-2018}. Short-term changes are intensified nowadays thanks to the ubiquity of the Web, the intense dynamics of social media and the dramatic increase in the speed of communication brought about by communication and Web technologies. These suggest that lexical changes can materialize in much shorter time frames than it was in the past.

\section{Summary, Conclusions and Research Directions}\label{sec:summary}
We summarize below the main observations of our survey. 

First of all, we note that the field has grown rapidly in the last few years resulting in a variety of techniques for lexical semantic change detection, ranging from counting approaches over generative models to neural network based word embeddings. The state of the art is represented by methods based on word embedding techniques. However, most of these approaches are sense-agnostic, effectively focusing on the mixture of word senses expressed by a lexeme. Although some claim that their methods utilize the dominant word sense, they use each occurrence of the lexeme or word form without detecting if it is indeed representing the dominant sense or not. 

Another common shortcoming is that only a few approaches propose techniques capable of analyzing semantic change in words with relatively few occurrences. The amount of data for low-frequency words may be insufficient to construct reliable hypotheses using standard methods. Dynamic embeddings seem to offer a more suitable alternative with respect to small datasets. 
When moving to sense-differentiated embeddings, even more data is likely needed, and the dynamic embeddings can be a path forward.
In relation to this, a common restriction of the discussed methods is that they work on a vocabulary common to all the investigated time periods and make use of the \emph{K} most common words. In some cases, the word frequencies are first normalized per year to avoid a dominance of modern words (since the available digital datasets grow in size over time). Still, this means that only words extant in the datasets over the entire time period contribute to the analysis, both in that they are the only words for which change can be detected, but also because they cannot contribute to the meaning of present words. A word like the Old Swedish legal term \emph{bakvaþi}, meaning `to accidentally stab someone standing behind you when taking aim to swing your sword forward', is only valid for a period and then disappears from our vocabulary. By ignoring this word, we will not capture any changes regarding the word, which has a very interesting story, but we also prevent it from contributing to the meaning of any of our other \textit{K} words.

In addition, since most of the corpora are not first normalized with respect to spelling variation, many common words are ignored only because their spelling has changed over time. For example, {\em infynyt, infinit, infinyte, infynit, infineit} are all historical spelling variations used at different times for the word now spelled {\em infinite} \cite{OED}. To properly address the problem of discovering and describing language change, we need to combine spelling variation, sense variation and lexical replacements in one framework.

Next, while a sense change may be successfully detected \emph{as a diachronic process}, determining the exact time point of semantic change requires the formulation of auxiliary hypotheses about the criteria to be used for determining this. Such criteria are obviously dependent on the available data. For most historical periods we have only texts typically produced by a small and skewed sample of the entire language community. Will thresholds of occurrence in historical texts faithfully reflect underlying change points? 

When it comes to evaluating methods and systems, there is a general lack of standardized evaluation practices. Different papers use different datasets and testset words, making it difficult or impossible to compare the proposed solutions. Proper evaluation metrics for semantic change detection and temporal analog detection have not been yet established. Furthermore, comparing methods proposed by different groups is difficult due to varying preprocessing details. For example, filtering out infrequent words can impact the results considerably and different papers employ different thresholds for removing rare words (e.g., some filter out words that appear less than 5 times, others less than 200 times). 

Only a few proposals seem to allow for automatically outputting evidence of change to explain to users the nuances of the sense change and to provide concrete examples. Change type determination by automatic means is one step towards this.  Related to this is the need for more user-friendly and extensive visualization approaches for diachronic conceptual change analysis given its inherent complexity. One should keep in mind that many researchers in, for example, the humanities will not accept tools that require programming skills on the part of the user, yet they require tools that are powerful enough for answering non-trivial questions and for enabling in-depth investigation.

The issue of interdependence between semantic changes of different words is also an interesting avenue of research. Most of the surveyed approaches focus on a single word, with only a few authors proposing to view sense change of a target word in relation to another reference word. Future approaches may take entire concepts or topics for investigation so that sense fluctuations of a given word would be seen in the context of changes of other words that may represent the same concept, the same topic or may be semantically related in some other way. Rather than analyzing diachronic conceptual change independently from the changes of other words, a more exhaustive approach could consider also senses of words belonging to an intricate net of word-to-word inter-relation. This could result in a more complete and accurate understanding of why and how a given word changed it sense.

Finally, we note that the linguistic study of semantic change has
traditionally been pursued in the context of single languages or
language families, and on limited data sets. In particular, nearly all the proposed approaches in the computational literature reviewed here are applied to English data only, due to the dominant position of English in various respects, which is reflected not least in the limited availability of datasets in other languages. Notably, the need for diachronic corpora in other languages than English has also been emphasized in the mentioned survey by \cite{tang-2018}.
Even if some ``laws'' of semantic change have been suggested \citep[\eg][]{wilkins,traugott-dasher}, and general classifications of semantic changes into types have been proposed \citep[see][]{urban-2015}, albeit also questioned \citep[see][]{fortson-2003}, the field is still underdeveloped with regard to its empirical basis. For example, it would be necessary to carefully consider whether the underlying corpus is indeed representative of the given language, and does not introduce any bias towards a particular region, gender, social group, and so on, before making any general claims. Approaches that rely on corroborating results using different datasets could be helpful here, especially if informed by a solid knowledge of linguistic methodology and applied to a significant number of genetically, typologically and geographically diverse languages, allowing for both extension and validation of databases such as the \emph{catalogue of semantic shifts} manually compiled by \citet{zalizniak-etal-2012}. How applicable the investigated methods will be to other languages is ultimately an empirical matter, but we see no reasons not to be optimistic in this regard.

In view of the above finding we list below several recommendations: 
     \begin{enumerate}
     \item When showing and discussing results in a paper, the authors should provide their viewpoint and justification thereof, whether these results are correct or not, and why. 
     \item Always use some sort of control, be it time-stable words or a control dataset, since in isolation, numbers are not sufficient.
     \item While there have been several methods proposed so far for automatically detecting semantic change, still there are no solutions for automatically generating the ``story of the word''. Such story-telling would help to concisely explain how the term changed, perhaps giving also a reason for the change (e.g., a new invention). Automatically detecting the type of change could be seen as the first step towards this goal.
     \end{enumerate}

\section*{Acknowledgements}

\starttwocolumn

\balance
\bibliography{SC-survey}

\onecolumn
\section*{Biographies of Authors}

\textbf{Nina Tahmasebi} is a researcher in Natural Language Processing at the University of Gothenburg, Sweden. She obtained a Ph.D. in Computer Science from the University of Hanover, Germany, in 2013. Her main research interest lie in automatic detection of diachronic language change, in particular semantic change, but she is interested in information extraction and change detection in general. She is the PI of the project Towards Computational Lexical Semantic Change Detection (https://languagechange.org/). She has done work in social media analysis, sentiment mining, summarization, and text mining for the digital humanities, and lately also on methodological issues resulting from applying data science to the digital humanities, to generate representative and sustainable humanities knowledge. 

\textbf{Lars Borin} is professor of Natural Language Processing at the University of Gothenburg, Sweden. 
He is also co-director of
Spr\aa{}kbanken (the Swedish Language Bank) and the national coordinator
of the Swedish activities in the European CLARIN ERIC research
infrastructure. His research interests include historical, areal and
typological linguistics, computational lexical semantics, and
methodological aspects of language technology, especially as applied
to low-density language varieties.

\textbf{Adam Jatowt} is an associate professor at Kyoto University, Japan. He has obtained Ph.D in Information Science in 2005 from the University of Tokyo. His research interests include broad aspects in text mining, temporal information retrieval from text, web archive search and mining and information comprehensibility. 

All three authors are co-organizers of the 1st International Workshop on Computational Approaches to Historical Language Change, to be held in conjunction with ACL2019.

\end{document}